\documentclass[journal]{IEEEtran}
\IEEEoverridecommandlockouts
\usepackage{cite}
\usepackage{graphicx}
\usepackage{amsmath,amssymb,amsfonts,url}
\usepackage{algorithm}
\usepackage{algorithmicx}
\usepackage{algpseudocode}
\usepackage{amsmath}
\usepackage[table*]{xcolor}
\xdefinecolor{gray95}{gray}{0.55}
\xdefinecolor{gray25}{gray}{0.8}
\usepackage{caption}

\usepackage{subfigure}
\usepackage{textcomp}
\usepackage{xcolor}
\usepackage{bm}
\usepackage{multirow}
\usepackage{booktabs}
\def\BibTeX{{\rm B\kern-.05em{\sc i\kern-.025em b}\kern-.08em
    T\kern-.1667em\lower.7ex\hbox{E}\kern-.125emX}}

\begin{document}
	
\title{Learning Adaptive Differential Evolution Algorithm from Optimization Experiences by Policy Gradient}
	
\author{Jianyong Sun,~\IEEEmembership{Senior~Member, IEEE}, Xin Liu, Thomas B\"{a}ck,~\IEEEmembership{Senior~Member, IEEE}, and Zongben Xu
	
\thanks{J. Sun, X. Liu and Z. Xu are with the School of Mathematics and Statistics, Xi'an Jiaotong University, Xi'an, China, 710049. email: jy.sun@xjtu.edu.cn, liuxin17@stu.xjtu.edu.cn, zbxu@mail.xjtu.edu.cn}\thanks{T. B\"{a}ck is with the Leiden Institute of Advanced Computer Science, Leiden University, Leiden, The Netherlands. email: t.h.w.baeck@liacs.leidenuniv.nl}}

\maketitle
	
\begin{abstract}
Differential evolution is one of the most prestigious population-based stochastic optimization algorithm for black-box problems. The performance of a differential evolution algorithm depends highly on its mutation and crossover strategy and associated control parameters. However, the determination process for the most suitable parameter setting is troublesome and time-consuming. Adaptive control parameter methods that can adapt to problem landscape and optimization environment are more preferable than fixed parameter settings. This paper proposes a novel adaptive parameter control approach based on learning from the optimization experiences over a set of problems. In the approach, the parameter control is modeled as a finite-horizon Markov decision process. A reinforcement learning algorithm, named policy gradient, is applied to learn an agent (i.e. parameter controller) that can provide the control parameters of a proposed differential evolution adaptively during the search procedure. The differential evolution algorithm based on the learned agent is compared against nine well-known evolutionary algorithms on the CEC'13 and CEC'17 test suites. Experimental results show that the proposed algorithm performs competitively against these compared algorithms on the test suites. 
\end{abstract}


\section{Introduction}\label{intro}

Among many evolutionary algorithm (EA) variants, differential evolution (DE)~\cite{Storn95,Storn97} is one of the most prestigious due to its exclusive advantages such as automatic adaptation, easy implementation, and very few control parameters~\cite{das16,neri10}. The DE variants have been successfully applied to a variety of real-world optimization problems~\cite{das11,das16}, and have been considered very competitive in the evolutionary computation community according to their performances on various competitions~\cite{Tanabe14,Brest19,skvorc19,Price19}. 

However, DE has also some drawbacks such as stagnation, premature convergence, sensitivity/insensitivity to control parameters and others~\cite{das11}. Although various factors, such as dimensionality of the decision space and the characteristics of the optimization problems, can result in these drawbacks, the bad choice of control parameters (namely, the scale factor $F$, the crossover rate $CR$ and the population size $N$) is one of the key problems~\cite{Eiben99,Karafotias15}.

It is well acknowledged that control parameters can significantly influence the performance of an evolutionary algorithm~\cite{Eiben99}, and this also holds for the control parameters of DE. In the early days of research in DE, the control parameters are usually set by trial-and-error~\cite{Roger02,Ronkkonen05,Tvrdik06} according to the {\em optimization experiences} gained from applying them to a set of test problems. Once the control parameters are set, they are fixed along the search procedure (this parameter determination approach is usually referred to as ``parameter tuning''). For examples, in~\cite{Kaelo07}, $F$ and $CR$ are suggested to be in the range of $[0.4, 1]$ and $[0.5, 0.7]$, respectively, while $N$ is suggested to be $[2-40]n$ where $n$ is the problem dimension~\cite{Ronkkonen05}. The trial-and-error approach is usually time-consuming, not reliable and inefficient~\cite{Tvrdik09}.

Along with the study of the exploration-exploitation relation in EA, it is found that the optimal control parameter setting for a DE algorithm is problem-specific, dependent on the state of the evolutionary search procedure, and on the different requirements of problems~\cite{Feoktistov06}. Further, different control parameters impose different influences on the algorithmic performance in terms of effectiveness, efficiency and robustness~\cite{Brest09}. Therefore, it is generally a very difficult task to properly determine the optimal control parameters for a balanced algorithmic performance due to various factors, such as problem characteristics and correlations among them.

Some researchers claim that $F$ and $CR$ both affect the convergence and robustness, but $F$ is more related to convergence~\cite{Price08}, $CR$ is more sensitive to the problem characteristics~\cite{Qin05}, and their optimal values correlate with $N$~\cite{Ilonen03}. $N$ could cause stagnation if it is too small, and slow convergence if it is too big~\cite{Feoktistov06}. Its setting strongly depends on problem characteristics (e.g. separability, multi-modality, and others).

It is also observed that the control parameters should be differently set at different generations simply due to changing requirements (i.e., exploration vs. exploitation) in different phases of the optimization run. For example, researchers generally believe that $F$ should be set larger at the beginning of the search to encourage exploration and smaller in the end to ensure convergence~\cite{Li10}. In light of this observation, parameter control methods, i.e., methods for changing the control parameter values dynamically during the run, have become one of the main streams in DE studies~\cite{dragoi16}. These methods are referred to as ``parameter control''.

In~\cite{Tanabe20}, the parameter control methods are classified as deterministic, adaptive, and self-adaptive, while a hybrid control is included in~\cite{dragoi16}. In this paper, we propose to classify the parameter control methods based on how they are set according to online information collected during the evolutionary search. Our classification differentiates whether parameters are learned during search or not, and is provided next.

\subsubsection{Parameter Control Without Learning} In this category, online information of any form is not reflected in the control of the parameters. Rather, a simple rule is applied deterministically or probabilistically along the search process.

The simple rule is constructed usually in three ways. First, no information is used at all. For example, in~\cite{Das05, Zou13}, $F$ is sampled uniformly at random from a prefixed range at each generation for each individual. In~\cite{Wang11}, a combination of $F$ and $CR$ is randomly picked from three pre-defined pairs of $F$ and $CR$ at each generation for each individual.

Second, the simple rule is time-varying depending on the generation. For example, in~\cite{Das05}, $F$ decreases linearly, while in~\cite{Draa15}, $F$ and $CR$ are determined based on a sinusoidal function of the generation index. In~\cite{Tanabe14}, the population size $N$ is linearly decreased with respect to the number of fitness evaluation used.

Third, information collected in the {\em current generation}, such as the range of the fitness values, the diversity of the population, the distribution of individuals and the rank of individuals, is used to specify the simple rule. For example, in~\cite{Ali04}, the minimum and maximum fitness values at current population are used to determine the value of $F$ at each generation. In~\cite{Tirronen09}, $F$ and $CR$ are sampled based on the diversity of the objective values of the current population, while the average of the current population in the objective space is used in~\cite{Jia09}. The individual's rank is used to determine $F$ and $CR$ for each individual in~\cite{Takahama12}. It is used to compute the mean values of the normal distributions associated with $F$ and $CR$ in~\cite{Tang15} from which $F$ and $CR$ are sampled for each individual.

\subsubsection{Parameter Control With Learning from the Search Process} In this category, collectable information during the search process is processed for updating the control parameters. The information used in these methods is mainly the successful trials obtained by using previous $F$ and $CR$ values.

There are mainly three ways. First, a trial $F$ and $CR$ is decided by an $\epsilon$-greedy strategy, as initially developed in~\cite{Brest06}. That is, if a uniformly sampled value is less than a hyper-parameter $\epsilon$, a random number in $[0,1]$ is uniformly sampled as the trial $F$ (resp. $CR$); otherwise previous $F$ (resp. $CR$) is used. If the trial by using the $F$ and $CR$ is successful, the sampled $F$ and $CR$ will be passed to the next generation; otherwise the previous $F$ and $CR$ will be kept. The sampled $F$ value is taken as a perturbation in~\cite{Lezama19}. The hyper-parameter $\epsilon$ is adaptively determined either by the current population diversity~\cite{Tirronen09}, or by fitness values~\cite{Jia09}.

Second, successful control parameters are stored in memory (or pool) during the optimization process which are then used to create the next parameters~\cite{Qin09,Zhang09}. For example, the median (or mean) value of the memory values is used as the mean of the control parameter distribution for sampling new control parameters~\cite{Qin09}. The distribution is assumed to be normal~\cite{Qin09} or Cauchy~\cite{Yang08b}. To make the sampling adapt to the search state, hyper-parameters are proposed in the distribution~\cite{Tvrdik06,Yang08b} while the hyper-parameters are updated at each generation according to the success of previous distributions.

Third, some authors proposed to update the mean of the control parameter's distribution through a convex linear combination between the arithmetic~\cite{Zhang09} or Lehmer~\cite{Tanabe14,Tanabe13} mean of the stored pool of successful and the current control parameters. In~\cite{Awad2017Ensemble}, an ensemble of two sinusoidal waves is added to adapt $F$ based on successful performance of previous generations. In~\cite{Mohamed2017LSHADE}, a semi-parameter adaptation approach based on randomization and adaptation is developed to effectively adapt $F$ values. In~\cite{Zamuda19}, the control parameters are also modified based on the ratio of the current number of function evaluations to the maximum allowed number of function evaluations. A grouping strategy with an adaptation scheme is proposed in~\cite{MENG201980,MENG201892} to tackle the improper updating method of $CR$.  In~\cite{Brest16,ZHAO201630}, a memory update mechanism is further developed, while new control parameters are assumed to follow normal and Cauchy distribution, respectively. In~\cite{PSOupdate}, the control parameters are updated according to the formulae of particle swarm optimization, in which the control parameters are considered as particles and evolved along the evolution process.

In this paper, we propose a novel approach to adaptively control $F$ and $CR$. In our approach, the control parameters at each generation are the output of a non-linear function which is modeled by a deep neural network (DNN). The DNN works as the parameter controller. The parameters of the network are learned from the experiences of optimizing a set of training functions by a proposed DE. The learning is based on the formalization of the evolutionary procedure as a finite-horizon Markov decision process (MDP). One of the reinforcement learning algorithms, policy gradient~\cite{rlbook}, is applied to optimize for the optimal parameters of the DNN.

This method can be considered as an automatic alternative to the trial-and-error approach in the early days of DE study. Note that the trial-and-error approach can only provide possible values or ranges of the control parameters. For a new test problem, these values need to be further adjusted which could result in spending a large amount of computational resources. Our approach does not need such extra adjustment for a new test problem. It can provide control parameter settings not only adaptively to the search procedure, but also to test problems. 

In the remainder of the paper, we first introduce deep learning and reinforcement learning which are the preliminaries for our approach in Section~\ref{dl}. Section~\ref{l2l2} presents the proposed learning method. Experimental results are presented in Section~\ref{res} and~\ref{sen} including the comparison with several well-known DEs and a state-of-the-art EA. The related work is presented in Section~\ref{rw}. Section~\ref{con} concludes the paper.

\section{Parameterized Knowledge Representation by Deep and Reinforcement Learning}\label{dl}

\subsection{Deep Learning}

Deep learning is a class of machine learning algorithms for learning data representation~\cite{dlbook}. It consists of multiple layers of nonlinear processing units for extraction of meaningful features. The deep learning architecture can be seen as a parameterized model for knowledge representation and a tool for knowledge extraction. It can also be seen as an efficient expert system. That is, given the current state of a system, through the deep neural network, a proper decision can be made if the deep network is optimally trained. The deep neural network has a high order of modeling freedom due to its large number of parameters, which make it able to make accurate predictions. Deep learning has had remarkable success in recent years in image processing, natural language processing and other application domains~\cite{dlbook}.

\subsection{Reinforcement Learning}\label{rls}

Reinforcement learning (RL) deals with the situation that an agent interacts with its surrounding environment. The aim of learning is to find an optimal policy for the agent that can respond to the environment for a maximized cumulative reward. RL can be modeled as a 4-tuple $(s, a, r, p_a)$ Markov decision process (MDP), where $s$ (resp. $a$, $r$ and $p_a$) represents state (resp. action, reward, and transition probability). 

Formally, at each time step $t$, a state $s_t$ and reward $r_t$ are associated to it. A policy $\pi$ is a conditional probability distribution, i.e. $\pi = p(A_t|S_t;\theta)$ with parameter $\theta$ where $A_t$ (resp. $S_t$) represents the random variable for action (resp. state). Given the current state $s_t$, the agent takes an action $a_t$ by sampling from $\pi$. Given this action, the environment responds with a new state $s_{t+1}$ and a reward $r_{t+1}$. The expectation of total rewards $U(\theta) = \mathbb{E}_{\tau \sim q(\tau)}\left(\sum_{t=0}^T \gamma_t r_t \right) $, where $\gamma_t$ denotes the time-step dependent weighting factor, is to be maximized for an optimal policy $\pi^*$ where the expectation is taken over trajectory $\tau = \{s_0, a_0,s_1,a_1,\cdots,a_{T-1},s_T, \cdots\}$ with the joint probability distribution $q(\tau)$.

To train a RL agent for the optimal parameter, various RL algorithms, such as temporal difference, Q-learning, SARSA, policy gradient and others have been widely used for different scenarios (e.g. discrete or continuous action and state space)~\cite{rlbook}. If the policy is represented by a deep neural network, it leads to the so-called deep RL, which has gained incredible success on playing games, such as AlphaGo~\cite{silver16}.

\section{Adaptive Parameter Control via Policy Gradient}\label{l2l2}

In this section, we show how to learn to control the adaptive settings of a typical DE.  Before presenting the algorithm, the notations used in the paper are listed in Table~\ref{notations}.

\begin{table}
	\caption{Notations. }\label{notations}
	\centering\begin{tabular}{l|l} \hline\hline
		$N$				& the population size \\
		$n$				& dimension of the decision variable\\
		$f$				& the objective (fitness) function \\ \hline
		$\mathbf{x}_i^t \in \mathbb{R}^n$	& the $i$-th individual at the $t$-th generation \\
		${\cal P}^t\in \mathbb{R}^{N\times n}$		& the $t$-th population \\
		${\cal F}^t \in \mathbb{R}^{N}$		& the fitness values of the $t$-th population \\
		${\cal V}^t \in \mathbb{R}^{N\times n}$		& the mutated population at the $t$-th generation \\
		$\widetilde{\cal P}^t \in \mathbb{R}^{N\times n}$ & the trial solutions obtained at the $t$-th generation \\
		$\widetilde{\cal F}^t\in \mathbb{R}^{N}$& the fitness values of the trial solutions \\ \hline
		${\cal U}_t $		& the statistics at the $t$-th generation \\
		${\cal H}_t$		& the memory at the $t$-th generation \\\hline
		\textbf{M}			& the mutation operator \\
		\textbf{CR}		& the crossover operator\\
		\textbf{S}			& the selection operator \\ \hline\hline

	\end{tabular}
\end{table}

\subsection{The Typical DE} \label{cur2pbest}

In the proposed DE, the current-to-$p$best/1 mutation operator and the binomial crossover are employed. In the current-to-$p$best/1 mutation operator~\cite{Zhang09}, at the $t$-th generation, for each individual $\mathbf{x}_i^t$, a mutated individual $\mathbf{v}_i^t$ is generated in the following manner:
\begin{equation}
\mathbf{v}_i^t=\mathbf{x}_i^t+F_i^t\cdot(\mathbf{x}_{\text{pbest}}^t-\mathbf{x}_i^t)+F_i^t\cdot(\mathbf{x}_{r_1}^t-\mathbf{x}_{r_2}^t), \label{eq_pbest}
\end{equation}where $\mathbf{x}_{\text{pbest}}^t$ 
is an individual randomly selected from the best $N \times p$ $(p\in(0,1])$ individuals at generation $t$. The indices $r_1$, $r_2$ are randomly selected from $[1,N]$ such that they differ from each other and $i$. 

The binomial crossover operator works on the target individual $\mathbf{x}_i^t=(x_{i,1}^t,..., x_{i,n}^t)$ and the corresponding mutated individual $\mathbf{v}_i^t=(v_{i,1}^t,..., v_{i,n}^t)$ to obtain a trial individual $\widetilde{\mathbf{x}}_i^t=(\widetilde{x}_{i,1}^t,..., \widetilde{x}_{i,n}^t)$ element by element as follows: 
\begin{equation}
\widetilde{x}_{i,j}^t =\begin{cases}
v_{i,j}^t, & \mbox{if} \; \text{rand}[0,1] \leq CR_i^t \ \text{or} \ j=j_{\text{rand}} \\
x_{i,j}^t, & \mbox{otherwise}
\end{cases}  \label{crossover}
\end{equation}where rand$[0,1]$ denotes a uniformly sampled number from $[0, 1]$ and $j_{\text{rand}}$ is an integer randomly chosen from $[1,n]$.

Given the trial individuals, the next generation is selected individual by individual. For each individual, if $f(\widetilde{\mathbf{x}}_i^t) \leq f(\mathbf{x}_i^t)$, then $\mathbf{x}_i^{t+1} = \widetilde{\mathbf{x}}_i^t$; otherwise $\mathbf{x}_i^{t+1} = \mathbf{x}_i^t$.

It is clear that the control parameters of the proposed DE include $N$ and $\{F_i^t, CR_i^t, 1\leq i\leq N\}$ at each generation. In the following, we will show how $\{F_i^t, CR_i^t, 1\leq i\leq N\}$  can be learned from optimization experiences.

\subsection{Embed Recurrent Neural Network within the Typical DE}

The evolution procedure of the proposed DE can be formalized as follows. At generation $t$, a mutation population ${\cal V}^t =\{\mathbf{v}_1^t, \cdots, \mathbf{v}_N^t\}$ is first generated by applying the mutation operator (denoted as $\mathbf{M}$) on the current population ${\cal P}^t$; a trial population $\widetilde{\cal P}^t = \{\widetilde{\mathbf{x}}_1^t, \cdots, \widetilde{\mathbf{x}}_N^t\}$ is further obtained by applying the binomial crossover (denoted as $\mathbf{CR}$) operator. The new population is then formed by applying the selection (denoted as $\mathbf{S}$) operation. In the sequel, we denote $\Theta_{\text{F}}^t = \{F_i^t, 1\leq i \leq N\}$ and $\Theta^t_{\text{CR}} = \{CR_i^t, 1\leq i\leq N\}$.

\begin{figure}
	\centering
	\includegraphics[scale=0.28]{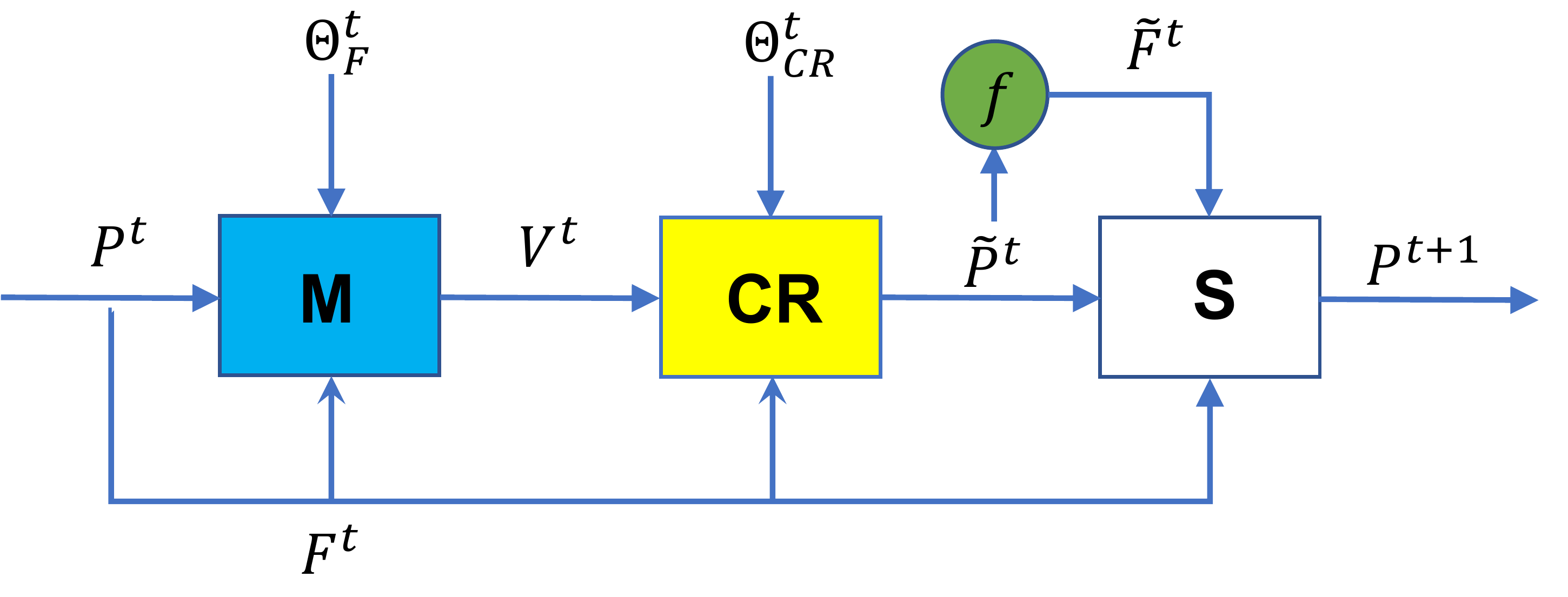}
	\caption{The flowchart of a typical DE at the $t$-th generation. In the figure, the mutation operator (resp. crossover and selection operator) is denoted as $\mathbf{M}$ (resp. $\mathbf{CR}$ and $\mathbf{S}$). }\label{flowchart0}
\end{figure}

Formally, the evolution procedure can be written as follows:
\begin{equation}\label{ea_procedure}
\begin{split}
{\cal V}^t 	& =\textbf{M}({\cal P}^t;\Theta_{\text{F}}^t);  \\
\widetilde{\cal P}^t 	& =\textbf{CR}({\cal V}^t, {\cal P}^t;\Theta_{\text{CR}}^t); \\
\widetilde{\cal F}^t	&=  f(\widetilde{\cal P}^t ); \\
{\cal P}^{t+1}, {\cal F}^{t+1}&=  \textbf{S}({\cal F}^t, \widetilde{\cal F}^t, {\cal P}^t, \widetilde{\cal P}^t).
\end{split}
\end{equation}
Due to the stochastic nature of the mutation and crossover operators, the evolution procedure can be considered as a stochastic time series. The creation of solutions at generation $t$ depends on information collected from previous generations from $1$ to $t-1$. Fig.~\ref{flowchart0} shows the flowchart of the procedure at the $t$-th generation.

As discussed in the introduction, recent studies focused on controlling $F$ and $CR$ by learning from online information. Various kinds of information are derived and used to update the control parameters for the current population. Generally speaking, the updated control parameters can be considered as output of a non-linear function with the collected information as input. Since an artificial neural network (ANN) is a universal function approximator~\cite{Cybenko89}, this motivates us to take an ANN to approximate the non-linear function. 

\begin{figure}[h]
	\centering
	\includegraphics[scale=0.25]{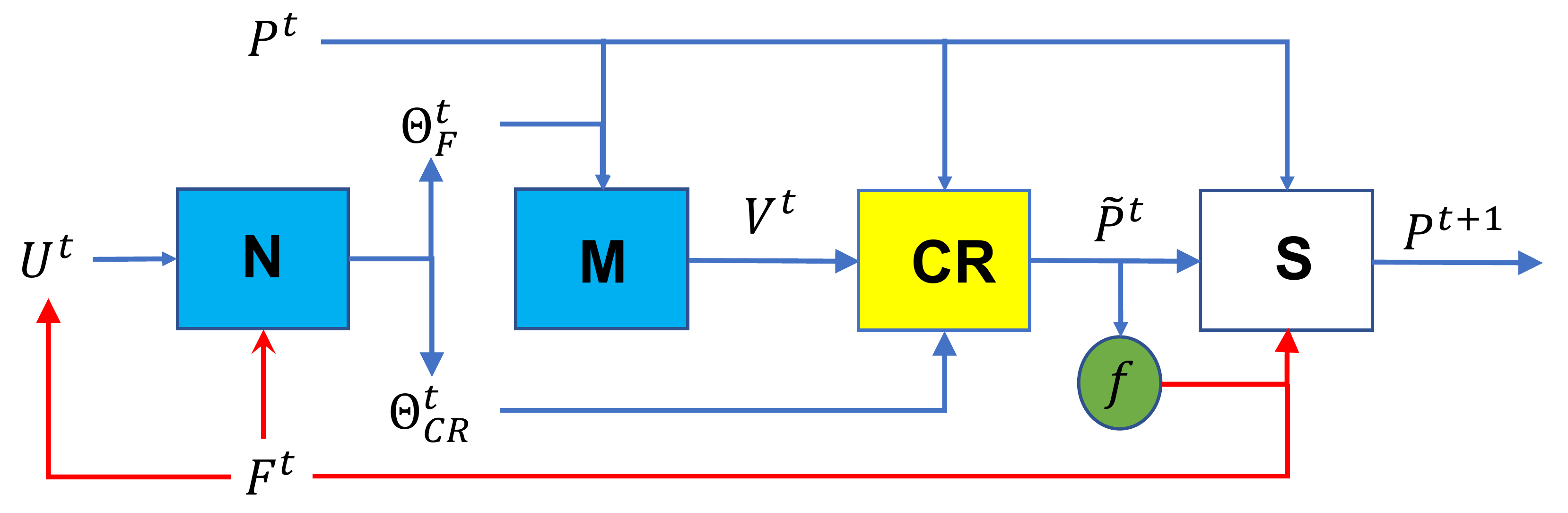}
	\caption{The proposed DE with embedded neural network (denoted as {\bf N}) at the $t$-th generation.}\label{flowchart}
\end{figure}

The neural network can be embedded in the evolution procedure as a parameter controller. Fig.~\ref{flowchart} shows the flowchart of the proposed DE with embedded neural network at the $t$-th generation. As illustrated in the figure, the neural network $\left({\bf N} ~\text{in the figure}\right)$ outputs $\Theta_{\text{F}}^t$ and $\Theta_{\text{CR}}^t$.  $\Theta_{\text{F}}^t$ is used as the input to the mutation operator, while $\Theta_{\text{CR}}^t$ is the input to the crossover operator. The mutated population (i.e. ${\cal V}^{t}$) and trial population (i.e. $\widetilde{\cal P}^{t}$) are then generated, respectively. ${\cal V}^{t}$ and ${\cal P}^{t}$ are the input to \textbf{CR}; $\widetilde{P}^{t}$ and ${\cal P}^{t}$ are the input to $\textbf{S}$ for selecting the new population ${\cal P}^{t+1}$. 

Existing research only utilizes the information collected from the current and/or previous generations for the parameter control. However, all the information until the current generation should have certain influences, although with different importances. The closer to the current generation, the more influential. 

\begin{figure}[htbp]
\centering
\includegraphics[scale=0.45]{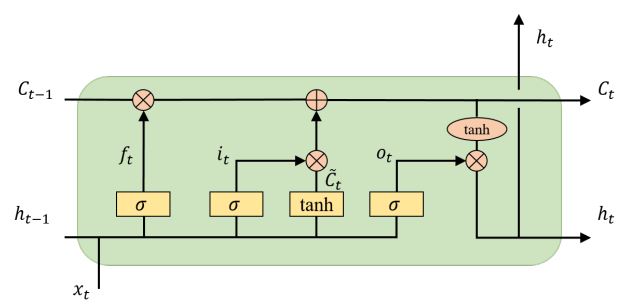}
\caption{The flowchart of the LSTM.}\label{lstm_flowchart}
\end{figure}

To accommodate the time-dependence feature, we take the neural network to be a long short-term memory (LSTM)~\cite{Hochreiter1997Long}. LSTM is a kind of recurrent neural network. As its name suggests, LSTM is capable of capturing long-term dependencies among input signals. There are a variety of LSTM variants. Its simplest form is formulated as follows:
\begin{eqnarray}
f_t &=& \sigma\left(W_f \cdot [h_{t-1}, x_t] + b_f\right) ;\nonumber\\
i_t &=& \sigma\left(W_i \cdot [h_{t-1}, x_t]+ b_i\right); \nonumber\\
\tilde{C}_t &=& \text{tanh}\left(W_c\cdot [h_{t-1}, x_t] + b_c\right); \nonumber\\
C_t &=& f_t \otimes C_{t-1} + i_t \otimes \tilde{C}_t; \nonumber\\
o_t &=& \sigma\left(W_o \cdot [h_{t-1}, x_t] + b_o \right); \nonumber\\
h_t &=& o_t \otimes \text{tanh}(C_t), \nonumber
\end{eqnarray}where $x_t$ is the input at step $t$, $[h_{t-1}, x_t]$ means the catenation of $h_{t-1}$ and $x_t$, $\otimes$ means Hadamard product, $\sigma$ and $\text{tanh}$ are the sigmoid activation function and tanh function, respectively:
\begin{equation}
\sigma(z) = \frac{1}{1+ e^{-z}}; \text{tanh}(z) = \frac{e^z - e^{-z}}{e^z + e^{-z}}. \nonumber
\end{equation}The parameters of the LSTM include $W_f,W_i, W_c$ and $W_o$ which are matrices and $b_f, b_i, b_c$ and $b_o$ which are biases. Fig.~\ref{lstm_flowchart} shows the flowchart of the LSTM. 
\vspace{-13pt}

Omitting the intermediate variables, the LSTM can be formally written as:
\begin{equation}
C_t, h_t = \text{LSTM}(x_t, h_{t-1}, C_{t-1}; \mathbf{W}),\label{lstm_formula}
\end{equation}where $\mathbf{W} = [W_f, W_i, W_c, W_o, b_f, b_i, b_c, b_o]$ denotes its parameter.

In our context, we consider the input to the LSTM as the catenation of ${\cal F}^t$ and ${\cal U}^t$ which is some statistics derived from ${\cal F}^t$, and denote ${\cal A}^t = [{\cal F}^t,{\cal U}^t]$\footnote{Here, $[{\cal F}^t,{\cal U}^t]$ also means to catenate the two vectors ${\cal F}^t$ and ${\cal U}^t$ into one single vector. }. In addition, we use ${\cal H}^t$ and ${\cal C}^t$ to represent the short and long term memory. Formally, the parameter controller can be written as follows:
\begin{eqnarray}\label{lstmall}
\begin{split}
 {\cal C}^t,{\cal H}^t  	&= {\text{LSTM}}({\cal A}^t, {\cal H}^{t-1}, {\cal C}^{t-1}; \mathbf{W}_L);\\
{\Theta}_{\text{F}}^t  		&= \text{FullConnect}({\cal H}^t; \mathbf{W}_F, b_F);\\
{\Theta}_{\text{CR}}^t	&= \text{FullConnect}({\cal H}^t;\mathbf{W}_C,b_C),
\end{split}
\end{eqnarray}where $\mathbf{W}_L$ is the parameter of LSTM, $\text{FullConnect}(\cdot; \mathbf{W},b)$ represents a fully-connected neural network with weight matrix $\mathbf{W}$ and bias $b$. Here 
\begin{eqnarray}\label{l_fcr}
\begin{split}
{\Theta}_{\text{F}}^t &= \sigma({{\cal H}^t}^\top \mathbf{W}_F +b_F); \\
{\Theta}_{\text{CR}}^t &= \sigma({{\cal H}^t}^\top\mathbf{W}_C + b_C).
\end{split}
\end{eqnarray}

In the sequel, we denote ${\Theta}^t = [\Theta_{\text{F}}^t, \Theta_{\text{CR}}^t]$ and use the following concise formula to represent Eq.~\ref{lstmall}: 
\begin{equation}
{\Theta}^t, {\cal C}^t,  {\cal H}^t = \text{LSTM}({\cal A}^t, {\cal H}^{t-1}, {\cal C}^{t-1}; \mathbf{W}),
\end{equation}where $\mathbf{W} = [\mathbf{W}_L, \mathbf{W}_F, b_F, \mathbf{W}_C, b_C].$ 

\subsection{Model the Evolution Search Procedure as an MDP}

To learn the parameters of the LSTM, i.e. the agent or the controller, embedded in the DE, we first model the evolution procedure of the proposed DE as an MDP with the following definitions of environment, state, action, policy, reward, and transition.

\subsubsection{Environment}  For parameter control, an optimal controller is expected to be learned from optimization experiences obtained when optimizing a set of optimization problems. Therefore, the environment consists of a set of optimization problems (called training functions). They are used to evaluate the performance of the controller when learning. Note that these training functions should have some common characteristics for which can be learned for a good parameter controller.

\subsubsection{State ($S_t$)} We take the fitness $\mathcal{F}^t$, the statistics ${\cal U}^t$ and the memories ${\cal H}^{t}$ as the state $S_t$.

Particularly, ${\cal U}^t$ includes the histogram\footnote{A histogram is constructed by dividing the entire range of values into a series of intervals (i.e. bins), and count how many values fall into each bin.} of the normalized ${\cal F}^t$ (denoted as $\mathbf{h}_t$) and the moving average of the histogram vectors over the past $g$ generations (denoted as $\bar{\mathbf{h}}_t$). Formally, \begin{eqnarray}\label{ustat}
\begin{split}
\bar{f}_i 			&= \frac{f_i - \min{\{\cal F}^t\}}{\max{\{\cal F}^t\} -  \min{\{\cal F}^t\}};\\
\mathbf{h}_t		&= \text{histogram}(\{\bar{f}_i\}, b); \\
\bar{\mathbf{h}}_t	&= \frac{1}{g} \sum_{i = t-\text{g}}^{t-1} \mathbf{h}_i ,
\end{split}
\end{eqnarray}where $b$ is the number of bins. The lower (resp. upper) range of the bins is defined as $\min \{\mathcal{F}_t\}$ (resp. $\max \{\mathcal{F}_t\}$). That is, to derive ${\cal U}^t$, the fitness values of the current population are first normalized. Its histogram is then computed and taken as the input to LSTM. This is to represent the information of the current population. Further, the statistics represented by $\bar{\mathbf{h}}_t$ is computed as the information from previous search history. 

It should be noted that the statistics ${\cal U}^t$ is computed at each generation w.r.t. the current population, not to each individual.

\subsubsection{Action ($A_t$) and Policy ($\pi$)}

In the MDP, given state $S_t$, the agent can choose (sample) an action from policy $\pi$ defined as a probability distribution $p(A_t|S_t;\theta)$ where $\theta$ represents the parameters of the policy. Here we define $A_t$ as the control parameters, i.e. $A_t = \{F_i^t, CR_i^t, 1\leq i\leq N\} \in \mathbb{R}^{2N}$.

Since the control parameters take continuous values, we assume the policy is normal. That is,
\begin{equation}
\begin{aligned}
&\pi(A_t|S_t) = {\cal N}(A_t|\text{LSTM}(S_t), \sigma^2) \\
&= \frac{1}{(2\pi\sigma^2)^N}\exp\left\{-\frac{1}{2\sigma^2} \left(A_t - \text{LSTM}(S_t;\mathbf{W})\right)^2\right\}. 
\end{aligned}
\end{equation}It is seen that the policy is uniquely determined by the LSTM parameter $\mathbf{W}$. 

\subsubsection{Reward ($r_{t+1}$)} The environment responds with a reward $r_{t+1}$ after the action. In our case, the reward $r_{t+1}$ is defined as the relative improvement of the best fitness
\begin{equation}
r_{t+1} = \frac{\max\{{\cal F}^{t+1} \} - \max\{ {\cal F}^{t}\} }{\max\{ {\cal F}^{t}\}} \label{r_eq}
\end{equation}where $\max\{ {\cal F}^{t} \}$ denotes the best fitness obtained at generation $t$. That is, after determining the control parameters, the mutation, crossover and selection operations are performed to obtain the next generation. The relative improvement is considered as the outcome of the application of the sampled control parameters.

A higher reward (improvement) indicates that the determined control parameters have a more positive impact on the search for global optimum.

\subsubsection{Transition} The transition is also a probability $p(S_{t+1}|A_t = a_t, S_t = s_t)$. In our case, the probability distribution is not available. It will be seen in the following that the transition distribution does not affect the learning.

\subsection{Learn the Control Parameter by Policy Gradient}

A variant of the RL algorithm, policy gradient (PG), is able to deal with the scenario when the transition probability is not known. PG works by updating the policy parameters via stochastic gradient ascent on the expectation of the reward
\begin{equation}
\theta_{t+1}=\theta_t + \alpha_t \nabla_{\theta} U(\theta_t), \label{theta_nn}
\end{equation}
where $\alpha_t$ denotes the learning rate and $\nabla_{\theta} U(\theta) $ is the gradient of the cumulative reward $U(\theta)$.

An evolutionary search procedure with a finite number of generations (denoted as $T$) can be considered as a finite-horizon MDP. For such an MDP, given previously defined state and action, a trajectory $\tau$ is $\{S_0, A_0, r_1,  \dots, S_{T-1}, A_{T-1}, r_T\}$. The joint probability of the trajectory can be written as\begin{equation}
q(\tau;\theta)= p(S_0)\prod_{t=0}^{T-1}\pi(A_t|S_t;\theta)p(S_{t+1}|A_t,S_t).\label{tau_density}
\end{equation}Further, $\nabla_{\theta} U(\theta)$ can be derived as follows: 
\begin{eqnarray}\label{policy_gradient}
\nabla_{\theta} U(\theta) &=&  \sum_{\tau} r(\tau) \nabla_{\theta} q(\tau;\theta) =  \sum_{\tau}r(\tau) \frac{\nabla_{\theta} q(\tau;\theta)}{q(\tau;\theta)} q(\tau;\theta) \nonumber\\
&=&  \sum_{\tau}r(\tau)q(\tau;\theta)  \nabla_{\theta}\left[ \log q(\tau;\theta) \right] \nonumber\\
&=&  \sum_{\tau} r(\tau) q(\tau;\theta)  \left[\sum_{t=0}^{T-1}\nabla_{\theta} \log \pi(A_t|S_t;\theta)\right]
\end{eqnarray}where $r(\tau)$ is the cumulative reward of the trajectory $\tau$. The expectation of Eq.~\ref{policy_gradient} can be calculated by sampling $L$ trajectories $\tau^1, \cdots, \tau^L$,
\begin{equation}
\nabla U_{\theta}(\theta)\approx\frac{1}{L}\sum_{i=1}^{L}r(\tau^i)\sum_{t=0}^{T-1}\nabla_{\theta} \ln \pi(A_t^{(i)}=a_t^{(i)}|S_t^{(i)}= s_t^{(i)};\theta) \label{policy_gradient_sampling}
\end{equation}where $a_t^{(i)}$ (resp. $s_t^{(i)}$) denotes action (resp. state) value at time $t$ in the $i$-th trajectory.

A detailed description on how to update $\mathbf{W}$ is given in Alg.~\ref{train_PG}. In the algorithm, to obtain the optimal $\mathbf{W}$, the optimization experience of a set of $M$ training functions is used. For each function, first a set of $L$ trajectories is sampled by applying the proposed DE (line~\ref{tj1}-\ref{tj2}) for $T$ generations. The control parameters of the proposed DE are obtained by forward computation of the LSTM given the present $\mathbf{W}$ at each generation. With the sampled trajectories, the reward at each generation for each optimization function is computed, and used for updating $\mathbf{W}$ (line~\ref{up}).  

\begin{algorithm*}[htbp]
	\caption{Learning to control the parameters of the DE} \label{train_PG}
	\begin{algorithmic}[1]
		\Require the LSTM parameter $\mathbf{W}$, the number of epochs $Q$, the number of training functions $M$, the population size $N$, the number of trajectories $L$,  the trajectory length $T$ and the learning rate $\alpha$
		\State Initialize $\mathbf{W}$ uniformly at random;
		\For{$epoch=1 \to Q$}
		\State Initialize ${\cal P}^0 =[\mathbf{x}_1^0, \cdots, \mathbf{x}_N^0] $ uniformly at random;
		\For{$k=1 \to M$}
		\State Set ${\cal{P}}_k^0 = {\cal P}^0$;
		\State Evaluate ${\cal F}_k^0 = \{f_k(\mathbf{x}_i^0), 1\leq i \leq N\}$; 
		\State \textcolor{blue}{$\triangleright${\em \ trajectory sampling;}}
		\For{$l=1 \to L$} \label{tj1}
		\State Set $t\leftarrow 0$, ${\cal H}_k^0 = \mathbf{0}$ and ${\cal C}_k^0 = \mathbf{0}$;
		\Repeat
		\State Compute ${\cal U}_k^t = [\mathbf{h}_t,\bar{\mathbf{h}}_t]$ by Eq.~\ref{ustat}; 
		\State Set ${\cal A}_k^t = [{\cal U}_k^t,{\cal F}_k^t]$;
		\State Apply LSTM: $\Theta_k^{t}, {{\cal C}_k^{t+1}, \cal H}_k^{t+1} \leftarrow \text{LSTM}({\cal A}_k^t, {\cal H}_k^{t}, {\cal C}_k^t;\mathbf{W});$
		\State Create the trial population: $\widetilde{\cal P}_k^t = [\mathbf{\widetilde{x}}_1^t,  \cdots, \mathbf{\widetilde{x}}_N^t]  \leftarrow \textbf{CR}\circ \textbf{M}({\cal P}_k^t;\Theta_k^{t});$
		\State Evaluate the trial population: $\widetilde{\cal F}_k^t \leftarrow \left\{f(\mathbf{\widetilde{x}}_i^t), 1\leq i\leq N\right\}$; 
		\State Form the new population: ${\cal P}_k^{t+1}, {\cal F}_k^{t+1} \leftarrow \textbf{S}({\cal F}_k^t, \widetilde{\cal F}_k^t, {\cal P}_k^t, \widetilde{\cal P}_k^t)$;
		\State Calculate $r_k^{t+1} $ using Eq.(\ref{r_eq}); and set $t \gets t+1$;
		\Until{$t\geq T$}
		\EndFor \label{tj2}		
		\EndFor
		\State \textcolor{blue}{$\triangleright${\em \ policy parameter updating;}}
		\State Update $\mathbf{W}$ using Eq.(\ref{policy_gradient_sampling}) and Eq.(\ref{theta_nn});\label{up} 
		\EndFor
		\State \Return {$\mathbf{W}$;}
	\end{algorithmic}
\end{algorithm*}

Note that in the learning, a set of optimization functions are used. For each optimization function, the proposed DE is applied for $T$ generations to sample the trajectories. Each trajectory can be considered as an optimization experience for a particular training function. For each function, there are $L$ trajectories sampled. $M$ functions can provide $L\times M$ optimization experiences. Learning from these experiences could thus be able to lead to a good parameter controller.

\subsection{Embed the Learned Controller within the DE}

After training, it is assumed that we have secured the required knowledge for generating control parameters through the learning from optimization experiences. Given a new test problem, the proposed DE with the learned parameter controller can be applied directly. The detailed algorithm, named as the learned DE (dubbed as LDE), is summarized in Alg.~\ref{test_PG}.

In Alg.~\ref{test_PG}, the evolution procedure is the same as a typical DE except that the control parameters are the samples of the output of the controller (line~\ref{lde3}-\ref{lde4}). One of the inputs of the LSTM, the statistics ${\cal U}^t$ is computed at each generation (line~\ref{lde2}), and the hidden information ${\cal H}^t$ and ${\cal C}^t$ are initialized (line~\ref{lde1}) and maintained during the evolution. 

Note that in the LDE, the controller contains knowledge learned from optimization experiences which can be considered as extraneous/offline information, while the use of ${\cal U}^t$, ${\cal H}^t$ and ${\cal C}^t$ represent the information learned during the search procedure which is intraneous/online information. 
\begin{algorithm}[htbp]
\caption{The Learned DE (LDE) } \label{test_PG}
\begin{algorithmic}[1]
\Require the trained agent with parameter $\mathbf{W}$
	\State Initialize population ${\cal P}^0$ uniformly at random;
	\State Evaluate ${\cal F}^0 = f({\cal P}^0)$;
	\State Set $g \leftarrow 0$, ${\cal H}^g = 0$ and ${\cal C}^g = 0$; \label{lde1}
	\While {the termination criteria have not been met}
		\State Compute ${\cal U}^g = [\mathbf{h}_g, \bar{\mathbf{h}}_g]$ by Eq.~\ref{ustat}; 
		\State Set ${\cal A}^g = [{\cal U}^g,{\cal F}^g]$; \label{lde2}
		\State $g \leftarrow g+1$;
		\State $\Theta^{g-1},{\cal C}^{g}, {\cal H}^{g}\leftarrow \text{LSTM}({\cal A}^{g-1}, {\cal H}^{g-1}, {\cal C}^{g-1};\mathbf{W})$;\label{lde3}
		\State $\widetilde{\Theta}^{g-1} \sim {\cal N}(\Theta^{g-1},\sigma^2)$; 
		\State ${\cal P}^{g},{\cal F}^{g} \leftarrow \text{DE}({\cal P}^{g-1}, {\cal F}^{g-1};\widetilde{ \Theta}^{g-1})$; \label{lde4}
	\EndWhile
\State \Return  $\mathbf{x}^* = \arg\max f({\cal P}^g)$
	\end{algorithmic}
\end{algorithm}	
The time complexity for one generation of LDE is $O(H^2 + N\cdot H + N\cdot n)$, where $H$ denotes the number of neurons used in the hidden layer.

\section{Experimental Study}\label{res}

In this section, we first present the implementation details of both Alg.~\ref{train_PG} and Alg.~\ref{test_PG}. The training details and the comparison results against some known DEs and a state-of-the-art EA are presented afterwards.

Some or all functions in CEC'13~\cite{liang2013problem}\footnote{\url{https://www.ntu.edu.sg/home/EPNSugan/index_files/CEC2013/CEC2013.htm}} are used as the training functions. In the comparison study, functions that have not been used for training from CEC'13 or CEC'17~\cite{wu2016problem}\footnote{\url{https://github.com/P-N-Suganthan/CEC2017-BoundContrained}} are tested. The CEC'13 test suite consists of five unimodal functions $f_1-f_5$, 15 basic multimodal functions $f_6-f_{20}$ and eight composition functions $f_{21}-f_{28}$. The CEC'17 test suite includes two unimodal functions $F_1$ and $F_2$, seven simple multimodal functions $F_3-F_9$, ten hybrid functions $F_{10}-F_{19}$, and ten more complex composition functions $F_{20}-F_{29}$.

When training, the following settings were used, including the number of epochs $Q=150$, the population size $N=50$ when $n=10$ and $N=100$ when $n=30$ for the mutation strategy, the number of bins $b=5$, the number of previous generations $g=5$, the trajectory length $T=50$, the number of trajectories $L = 20$, and the learning rate $\alpha = 0.005$.

For the experimental comparison, the same criteria as explained in~\cite{liang2013problem} and~\cite{wu2016problem} are used. Each algorithm is executed 51 runs for each function. The algorithm terminates if the maximum number of objective function evaluations ($\text{MAXNFE}$) exceeds $n\times10^4$ or the difference between the function values of the found best solution and the optimal solution (also called the function error value) is smaller than $10^{-8}$.

The compared algorithms include the following:

\textbf{DE}~\cite{Roger02}: the original DE algorithm with DE/rand/1/bin mutation and binomial crossover.

\textbf{JADE}~\cite{Zhang09}: the classical adaptive DE method in which the DE/current-to-$p$best/1 mutation strategy was firstly proposed. JADE has two versions. One is with an external archive. The external archive is to aid the generation of offspring when mutation. The other is without the archive. In JADE, each $F$ (resp. $CR$) is generated for each individual by sampling from a Cauchy (resp. normal) distribution and the location parameter of the distribution is updated by the Lehmer (resp. arithmetic) mean of the successful $F$'s (resp. $CR$'s).

\textbf{jSO}~\cite{Brest17jso}\footnote{Code is available at \url{https://github.com/P-N-Suganthan/CEC2017}}: which ranked second and the best DE-based algorithm in the CEC'17 competition. As an elaborate variation of JADE, two scale factors are associated with the mutation strategy. They are different from each other and limited within various bounds along the evolution.

\textbf{CoBiDE}~\cite{WANG14cobide}\footnote{Code is available at \url{https://sites.google.com/view/pcmde/}}: in which the $F$ (resp. $CR$) values are generated from a bimodal distribution consisting of two Cauchy distributions. Trial vectors are formed in the Eigen coordinate system built by the eigenvectors of the covariance matrix of the top individuals.

\textbf{cDE}~\cite{Tvrdik06}\footnote{Code is available in the extended CD version of the original article at \url{https://www1.osu.cz/~tvrdik/wp-content/uploads/men05_CD.pdf}}: in which both $F$ and $CR$ are selected from a pre-defined pool. The selection probability is proportional to the corresponding number of the successful individuals obtained from previous generations.

\textbf{CoBiDE-PCM}~\cite{Tanabe20}\footnote{Code is available at \url{https://sites.google.com/view/pcmde/}} and \textbf{cDE-PCM}~\cite{Tanabe20}\footnote{Code is available at \url{https://sites.google.com/view/pcmde/}}: that were proposed in~\cite{Tanabe20} for studying the effect of parameter control management. These two algorithms are largely consistent with CoBiDE and cDE, but are equipped with different mutation and crossover operators. They are ranked first or second on the BBOB benchmark in~\cite{Tanabe20}.

\textbf{HSES}~\cite{Zhang18HSES}\footnote{Code is available at \url{https://github.com/P-N-Suganthan/CEC2018}}: the winner of the bound constraint competition of CEC'18\footnote{The test functions in CEC'18 are the same as those in CEC'17 except the function $F_2$ in CEC'17 is removed.}. HSES is a three-phase algorithm. A modified univariate sampling is employed in the first stage for good initial points, while CMA-ES~\cite{Hansen03} is used in the second stage, followed by another univariate sampling, for local refinement.

The parameters and hyper-parameters of these methods are kept the same as the settings in the original references in our experiments. Table~\ref{para} shows the detailed variation operators used and the parameter (hyper-parameter) settings for the compared algorithms. 

It should be noted that JADE, cDE and CoBiDE are not tested on the CEC'13 or CEC'17 test suites in the original references, but on 20 basic functions, six basic functions and CEC'05~\cite{cec2005}, respectively. The parameters of these algorithms are tuned manually by grid search based on the mean fitness values found over a number of independent runs for each function. It is expected that the parameter tuning procedures of these algorithms are time-consuming and computationally intensive. However, we should be frank that there is a possibility that these algorithms' performances could be improved if their parameters are tuned on the CEC'13 or CEC'17 test suites.
	
\begin{table*}[htbp]
	\caption{Detailed reproduction operators used and parameter and hyper-parameter settings of the compared algorithms.} \label{para}
	\begin{center}
		\begin{tabular}{c|ll|l|l}
			\hline
			\multirow{2}*{Alg.} &\multicolumn{2}{|c|}{Operations} &Control &Hyper- \\
			\cline{2-3}
			&Mutation &Crossover  &Parameters &parameters \\
			\hline
			DE &DE/rand/1 &bin &$N=5n, F=0.5, CR=0.8$ & NA \\ \hline
			JADE &DE/current-to-pbest/1 &bin &$N=30,100~\text{when} ~n=10,30$  &$N_{\text{archive}}=N,\mu_F=0.5$  \\ 
				&  & &  & $\mu_{CR}=0.5,  c=0.1, p=0.05$ \\ \hline
			jSO &DE/current-to-pbest-w/1 &bin &$N_{\text{init}}=25\log \left(n\right) \sqrt{n}, N_{\text{min}}=4$ & $N_{\text{archive}}=20, M_F=0.5, M_{CR}=0.8$  \\ 
			 & & &  & $ p_{\text{max}}=0.25, p_{\text{min}}=0.125, H=5$ \\ \hline
			CoBiDE 	&DE/rand/1 &bin, covariance  &$N=60$ &$pb=0.4, ps=0.5$ \\ 
					& 		  & matrix learning & & \\ \hline
			CoBiDE-PCM &DE/best/1 &bin &$N=5\times n$ & NA \\  \hline
			cDE &DE/rand/1, DE/best/2 & bin & $N=\max \left(20, 2n\right), F=\left\{0.5, 0.8, 1\right\}$ &$n_0=2, \delta = 1/45$ \\ 
				& 	& & $CR=\left\{0, 0.5, 1\right\}$ & \\ \hline
			cDE-PCM &DE/rand/1 ($n=10$) &bin &$N=5n, F=\left\{0.5, 0.8, 1\right\}$ &$n_0=2, \delta = 1/45$ \\ 
			 &DE/best/1 ($n=30$) &bin & $CR=\left\{0, 0.5, 1\right\}$ & \\ \hline
			HSES &\multicolumn{2}{|c|}{modified univariate sampling, CMA-ES} &$M=200$ &$N_1=100, N_2=160, cc=0.96$ \\
			& & & & $I=20, \lambda=3\ln n +80$ \\ \hline
	\end{tabular}
	\end{center}
\end{table*}

\subsection{The Comparison Results on the CEC'13 Test Suite}

In this experiment, we use the first 20 functions $f_1-f_{20}$ in CEC'13 as the training functions. The remaining eight functions $f_{21}-f_{28}$ are used for comparison. As in~\cite{wu2016problem}, the function error value is used as the metric, and recorded for each run. The mean and standard deviation of the error values obtained for each function over 51 runs are used for comparison. In the following, the experimental results are summarized in tables, in which the means, standard deviations and the Wilcoxon rank-sum hypothesis test results are included. The best (minimum) mean values are typeset in bold.

The Wilcoxon rank-sum hypothesis test is performed to test the significant differences between LDE and the compared algorithms. The test results are shown by using symbols $+$, $-$, and $\approx$ in the tables. The symbol $+$ (resp. $-$, and $\approx$) indicates that LDE performs significantly worse than (resp. better than and similar to) the compared algorithms at a significance level of 0.05. The results are summarized in the ``WR" column of the tables.

Tables~\ref{13_10D_MAXFE} and~\ref{13_30D_MAXFE} summarize the experimental results when terminating at $\text{MAXNFE}$, while Table~\ref{13_10D_T} shows the results when the algorithms terminate at the maximum number of generations $T=50$ which is the number of generations used for training the agent.

\begin{table*}[htbp]
	\caption{Means and standard deviations of function error values for comparison of LDE on the CEC'13 benchmark suite for $n=10$ at generation $T=50$. }
	\label{13_10D_T}
	\begin{center}
		\setlength{\tabcolsep}{1.5mm}{
		\begin{tabular}{c|cc|ccc|ccc|ccc|ccc}
			\hline
			\multirow{2}*{} &\multicolumn{2}{c|}{LDE} &\multicolumn{2}{c}{DE} & &\multicolumn{2}{c}{JADE w/o archive} & &\multicolumn{2}{c}{JADE with archive} & &\multicolumn{2}{c}{jSO} & \\ \cline{2-15}
			&Mean & Std. Dev. &Mean & Std. Dev. & WR & Mean & Std. Dev. & WR & Mean & Std. Dev. & WR & Mean & Std. Dev. & WR \\
			\hline
			$f_{21}$ &5.39E+02 &3.73E+01 &4.07E+02 &6.79E+00 &$+$ &\textbf{3.99E+02} &\textbf{1.37E+01} &$+$ &4.01E+02 &9.99E-01 &$+$ &4.01E+02 &1.50E-01 &$+$ \\
			$f_{22}$ &1.71E+03 &1.83E+02 &2.00E+03 &2.30E+02 &$-$ &\textbf{1.55E+03} &\textbf{2.11E+02} &$+$ &1.56E+03 &2.29E+02 &$+$ &1.78E+03 &1.85E+02 &$\approx$ \\
			$f_{23}$ &2.25E+03 &2.23E+02 &2.16E+03 &1.90E+02 &$+$ &2.26E+03 &2.03E+02 &$\approx$ &2.29E+03 &2.16E+02 &$\approx$ &\textbf{1.94E+03} &\textbf{2.03E+02} &$+$ \\
			$f_{24}$ &2.28E+02 &5.48E+00 &2.23E+02 &6.16E+00 &$+$ &2.23E+02 &3.46E+00 &$+$ &2.24E+02 &3.52E+00 &$+$ &\textbf{2.21E+02} &\textbf{4.01E+00} &$+$ \\
			$f_{25}$ &2.30E+02 &2.35E+00 &2.22E+02 &3.14E+00 &$+$ &2.22E+02 &3.72E+00 &$+$ &2.22E+02 &5.97E+00 &$+$ &\textbf{2.19E+02} &\textbf{2.49E+00} &$+$ \\
			$f_{26}$ &1.98E+02 &1.14E+01 &1.68E+02 &1.81E+01 &$+$ &1.80E+02 &2.88E+01 &$+$ &1.89E+02 &3.99E+01 &$+$ &\textbf{1.62E+02} &\textbf{1.74E+01} &$+$ \\
			$f_{27}$ &6.66E+02 &3.08E+01 &5.25E+02 &4.71E+01 &$+$ &4.89E+02 &7.35E+01 &$+$ &5.32E+02 &6.64E+01 &$+$ &\textbf{4.72E+02} &\textbf{2.51E+01} &$+$ \\
			$f_{28}$ &9.31E+02 &1.05E+02 &5.24E+02 &4.73E+01 &$+$ &4.16E+02 &8.88E+01 &$+$ &4.43E+02 &9.77E+01 &$+$ &\textbf{3.92E+02} &\textbf{1.61E+01} &$+$ \\
			\hline
			{} &\multicolumn{2}{c}{$+/\approx/-$} &\multicolumn{3}{|c|}{7/0/1} &\multicolumn{3}{|c|}{7/1/0} &\multicolumn{3}{|c|}{7/1/0} &\multicolumn{3}{c}{7/1/0} \\
			\hline
		\end{tabular}}
	\end{center}
\end{table*}

\begin{table*}
	\caption{Means and standard deviations of function error values for comparison of LDE on the CEC'13 benchmark suite for $n=10$ when MAXNFE has been reached.} \label{13_10D_MAXFE}
	\begin{center}
		\setlength{\tabcolsep}{1.5mm}{
		\begin{tabular}{c|cc|ccc|ccc|ccc|ccc}
			\hline
			\multirow{2}*{} &\multicolumn{2}{c|}{LDE} &\multicolumn{2}{c}{DE} & &\multicolumn{2}{c}{JADE w/o archive} & &\multicolumn{2}{c}{JADE with archive} & &\multicolumn{2}{c}{jSO} & \\ \cline{2-15}
			&Mean & Std. Dev. &Mean & Std. Dev. & WR & Mean & Std. Dev. & WR & Mean & Std. Dev. & WR & Mean & Std. Dev. & WR \\
			\hline
			$f_{21}$ &\textbf{2.35E+02} &\textbf{9.04E+01} &3.75E+02 &7.10E+01 &$-$ &3.96E+02 &2.78E+01 &$-$ &4.00E+02 &5.68E-14 &$-$ &4.00E+02 &5.68E-14 &$-$ \\
			$f_{22}$ &1.73E+01 &2.60E+01 &4.83E+02 &2.82E+02 &$-$ &1.87E+01 &3.20E+01 &$-$ &2.68E+01 &4.00E+01 &$\approx$ &\textbf{6.80E+00} &\textbf{1.24E+01} &$+$ \\
			$f_{23}$ &7.66E+02 &2.62E+02 &1.21E+03 &1.51E+02 &$-$ &5.56E+02 &2.07E+02 &$+$ &5.88E+02 &2.29E+02 &$+$ &\textbf{2.13E+02} &\textbf{1.15E+02} &$+$ \\
			$f_{24}$ &\textbf{1.87E+02} &\textbf{3.80E+01} &1.97E+02 &1.96E+01 &$-$ &2.06E+02 &5.80E+00 &$\approx$ &2.06E+02 &5.39E+00 &$\approx$ &1.99E+02 &1.03E+01 &$-$ \\
			$f_{25}$ &2.01E+02 &7.87E+00 &2.00E+02 &1.41E+00 &$+$ &2.04E+02 &5.14E+00 &$\approx$ &2.05E+02 &1.08E+01 &$-$ &\textbf{2.00E+02} &\textbf{7.73E-05} &$+$ \\
			$f_{26}$ &1.16E+02 &1.80E+01 &1.34E+02 &2.92E+01 &$-$ &1.46E+02 &4.71E+01 &$\approx$ &1.57E+02 &6.06E+01 &$\approx$ &\textbf{1.05E+02} &\textbf{8.38E+00} &$+$ \\
			$f_{27}$ &3.31E+02 &4.85E+01 &\textbf{3.00E+02} &\textbf{7.96E-15} &$+$ &3.25E+02 &6.34E+01 &$+$ &3.63E+02 &9.75E+01 &$-$ &3.00E+02 &4.92E-05 &$+$ \\
			$f_{28}$ &\textbf{2.25E+02} &\textbf{9.67E+01} &2.88E+02 &4.71E+01 &$-$ &3.04E+02 &5.05E+01 &$\approx$ &3.00E+02 &3.43E+01 &$\approx$ &3.00E+02 &0.00E+00 &$-$ \\
			\hline
			{} &\multicolumn{2}{c}{$+/\approx/-$} &\multicolumn{3}{|c|}{2/0/6} &\multicolumn{3}{|c|}{2/4/2} &\multicolumn{3}{|c|}{1/4/3} &\multicolumn{3}{c}{5/0/3} \\
			\hline  
		\end{tabular}}
	\end{center}
\end{table*}

\begin{table*}
	\caption{Means and standard deviations of function error values for comparison of LDE on the CEC'13 benchmark suite for $n=30$ when MAXNFE has been reached.}\label{13_30D_MAXFE}
	\begin{center}
		\setlength{\tabcolsep}{1.5mm}{
		\begin{tabular}{c|cc|ccc|ccc|ccc|ccc}
			\hline
			\multirow{2}*{} &\multicolumn{2}{c|}{LDE} &\multicolumn{2}{c}{DE} & &\multicolumn{2}{c}{JADE w/o archive} & &\multicolumn{2}{c}{JADE with archive} & &\multicolumn{2}{c}{jSO} & \\ \cline{2-15}
			&Mean & Std. Dev. &Mean & Std. Dev. & WR & Mean & Std. Dev. & WR & Mean & Std. Dev. & WR & Mean & Std. Dev. & WR \\
			\hline
			$f_{21}$ &3.34E+02 &9.21E+01 &\textbf{2.70E+02} &\textbf{5.88E+01} &$+$ &2.91E+02 &6.54E+01 &$+$ &2.93E+02 &6.88E+01 &$+$ &3.12E+02 &8.28E+01 &$\approx$ \\
			$f_{22}$ &3.04E+02 &8.11E+01 &6.17E+03 &3.21E+02 &$-$ &9.40E+01 &3.07E+01 &$+$ &\textbf{8.58E+01} &\textbf{3.55E+01} &$+$ &1.22E+02 &4.88E+00 &$+$ \\
			$f_{23}$ &4.28E+03 &9.20E+02 &7.18E+03 &2.25E+02 &$-$ &3.37E+03 &3.70E+02 &$+$ &3.46E+03 &2.57E+02 &$+$ &\textbf{2.51E+03} &\textbf{3.14E+02} &$+$ \\
			$f_{24}$ &2.07E+02 &3.75E+00 &2.04E+02 &6.92E-01 &$+$ &2.06E+02 &4.97E+00 &$\approx$ &2.11E+02 &1.03E+01 &$\approx$ &\textbf{2.00E+02} &\textbf{1.06E-01} &$+$ \\
			$f_{25}$ &2.46E+02 &3.09E+01 &2.53E+02 &1.10E+01 &$\approx$ &2.75E+02 &1.08E+01 &$-$ &2.75E+02 &1.15E+01 &$-$ &\textbf{2.35E+02} &\textbf{1.56E+01} &$+$ \\
			$f_{26}$ &2.00E+02 &2.77E-03 &2.02E+02 &5.29E-01 &$-$ &2.18E+02 &4.55E+01 &$-$ &2.08E+02 &3.06E+01 &$-$ &\textbf{2.00E+02} &\textbf{2.16E-10} &$+$ \\
			$f_{27}$ &3.69E+02 &3.38E+01 &3.86E+02 &2.38E+01 &$-$ &6.23E+02 &2.07E+02 &$-$ &7.11E+02 &2.13E+02 &$-$ &\textbf{3.04E+02} &\textbf{3.11E+00} &$+$ \\
			$f_{28}$ &\textbf{3.00E+02} &\textbf{0.00E+00} &3.00E+02 &1.18E-06 &$-$ &\textbf{3.00E+02} &\textbf{0.00E+00} &$\approx$ &\textbf{3.00E+02} &\textbf{0.00E+00} &$\approx$ &\textbf{3.00E+02} &\textbf{0.00E+00} &$\approx$ \\
			\hline
			{} &\multicolumn{2}{c}{$+/\approx/-$} &\multicolumn{3}{|c|}{2/1/5} &\multicolumn{3}{|c|}{3/2/3} &\multicolumn{3}{|c|}{3/2/3} &\multicolumn{3}{c}{6/2/0} \\
			\hline
		\end{tabular}}
	\end{center}
\end{table*}

From Table~\ref{13_10D_T}, we can see that LDE does not perform satisfactorily. It is worse than the compared algorithms when the algorithm terminates at generation $T=50$. However, it performs better as the process of evolution continues up to MAXNFE. Note that in the training, the optimization procedure does not terminate at MAXNFE. The poor performance of LDE implies that the agent trained in $T=50$ generations is not good enough. However, the experimental results show that the knowledge learned in $T=50$ generations can be beneficial for the evolutionary search in further generations.

Tables~\ref{13_10D_MAXFE} and~\ref{13_30D_MAXFE} show that when MAXNFE has been reached, LDE exhibits superior overall performance as compared with DE/rand/1/bin and JADE with archive, and shows similar performance as compared with JADE without archive. However, LDE is outperformed by jSO in five and six out of eight 10-D and 30-D test functions, respectively. 

Specifically, on the eight 10-D complex composition functions, LDE performs better than DE on six functions $f_{21},f_{22},f_{23},f_{24},f_{26},f_{28}$, and surpasses JADE with archive on three functions $f_{21},f_{25},f_{27}$ and that without archive on two functions $f_{21},f_{22}$. However LDE performs statistically better than jSO on only three functions $f_{21},f_{24},f_{28}$ out of the eight test functions. For 30-D test problems, LDE yields better performance than DE on five functions $f_{22},f_{23},f_{26},f_{27},f_{28}$. LDE shows the same advantage over JADE with and without archive on $f_{25},f_{26},f_{27}$ on the CEC'13 benchmarks with 30-D. Again, LDE performs worse than jSO on six test functions $f_{22}-f_{27}$ with 30-D.

To see the overall performances, we rank the compared algorithms by using the average performance score (APS)~\cite{Bader11}. The APS is defined based on the error values obtained by the compared algorithms for the test functions. Suppose there are $m$ algorithms $A_1, \cdots, A_m$ to compare on a set of $M$ functions. For each $i,j \in [1,m]$, if $A_j$ performs better than $A_i$ on the $k$-th function $F_k, k \in [1, M]$ with statistical significance (i.e. $p < 0.05$), then set $\delta_{ij} = 1$ otherwise $\delta_{ij} = 0$. The performance score of $A_i$ on $F_k$ is computed as follows:
\begin{equation}
P_k(A_i) = \sum_{j \in [1, n]\backslash \{i\}} \delta_{ij}
\end{equation}The AP value of $A_i$ is the average of the performance score values of $A_i$ over the test functions. A smaller APS value indicates a better performance. 
\begin{table}[htbp]
\caption{The average ranks of the compared algorithms according to their APS values on the last eight functions in the CEC'13 test suite.}\label{ranking_13}
\centering
\setlength{\tabcolsep}{1.3mm}{
\begin{tabular}{cc|c|c|c|c|c}
	\hline
	\multicolumn{2}{c|}{\multirow{2}*{Alg.}} &\multirow{2}*{jSO}  &\multirow{2}*{LDE}   &JADE &JADE &\multirow{2}*{DE}     \\
	&  &  &  &w/o archive &with archive   &   \\
	\hline
	\multirow{3}*{} &$n=10$ &\textbf{0.625} &1.250 &1.375 &1.875 &1.625  \\
	&$n=30$ &\textbf{0.250} &1.625 &1.625 &1.875 &2.250   \\ 
	&Avg. &\textbf{0.4375} &1.437 &1.500 &1.875 &1.937 \\
	\hline
\end{tabular}}
\end{table}	
Table~\ref{ranking_13} summarizes the ranks of the compared algorithms in terms of their APS values. It can be seen that jSO is superior to LDE. LDE ranks the second, which is better than the other algorithms. This shows that the proposed method is quite promising.

\begin{figure}[htbp]
\centering
\subfigure[LDE]{\includegraphics[scale=0.18]{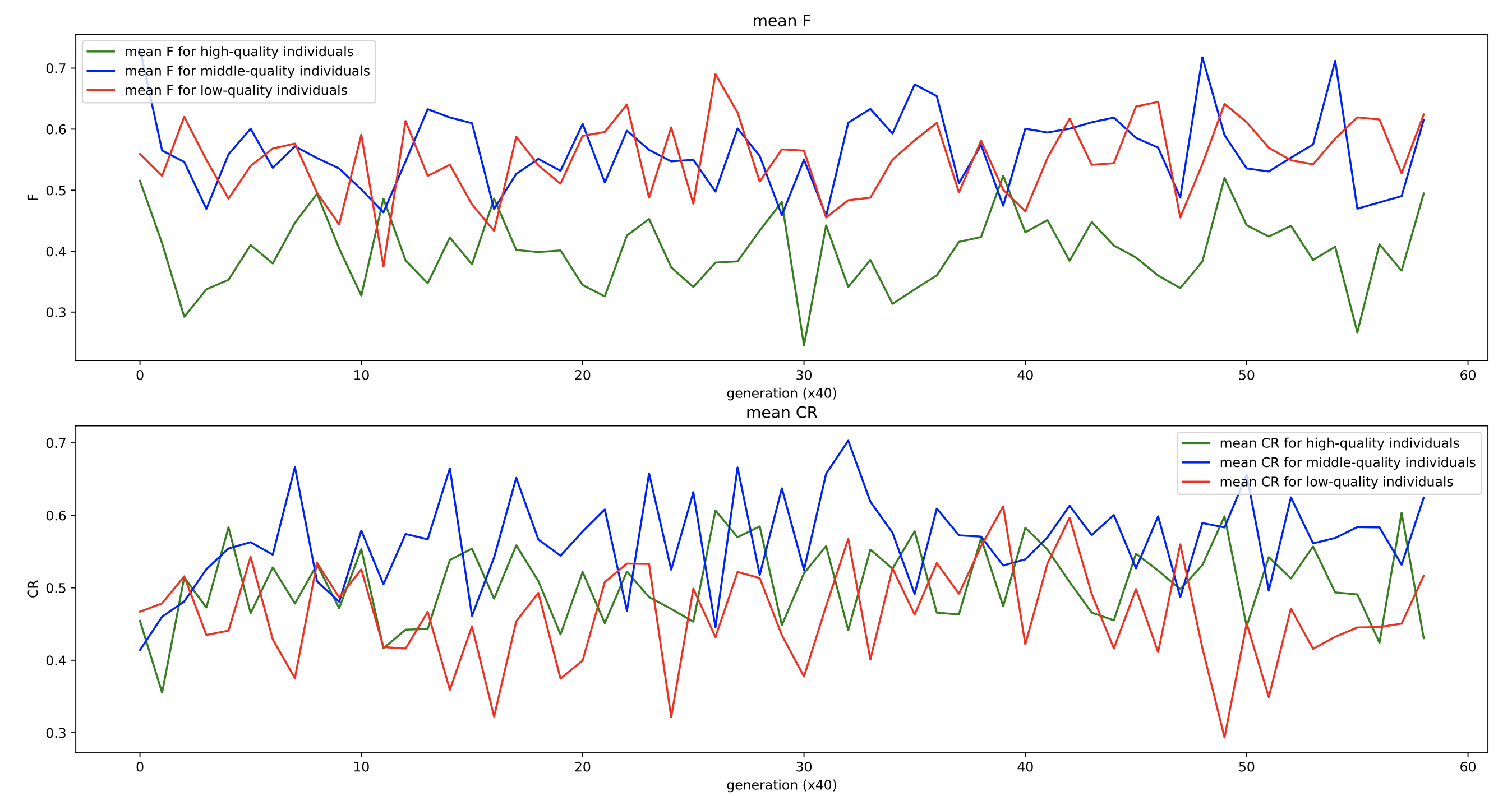}}
\subfigure[jSO]{\includegraphics[scale=0.1826]{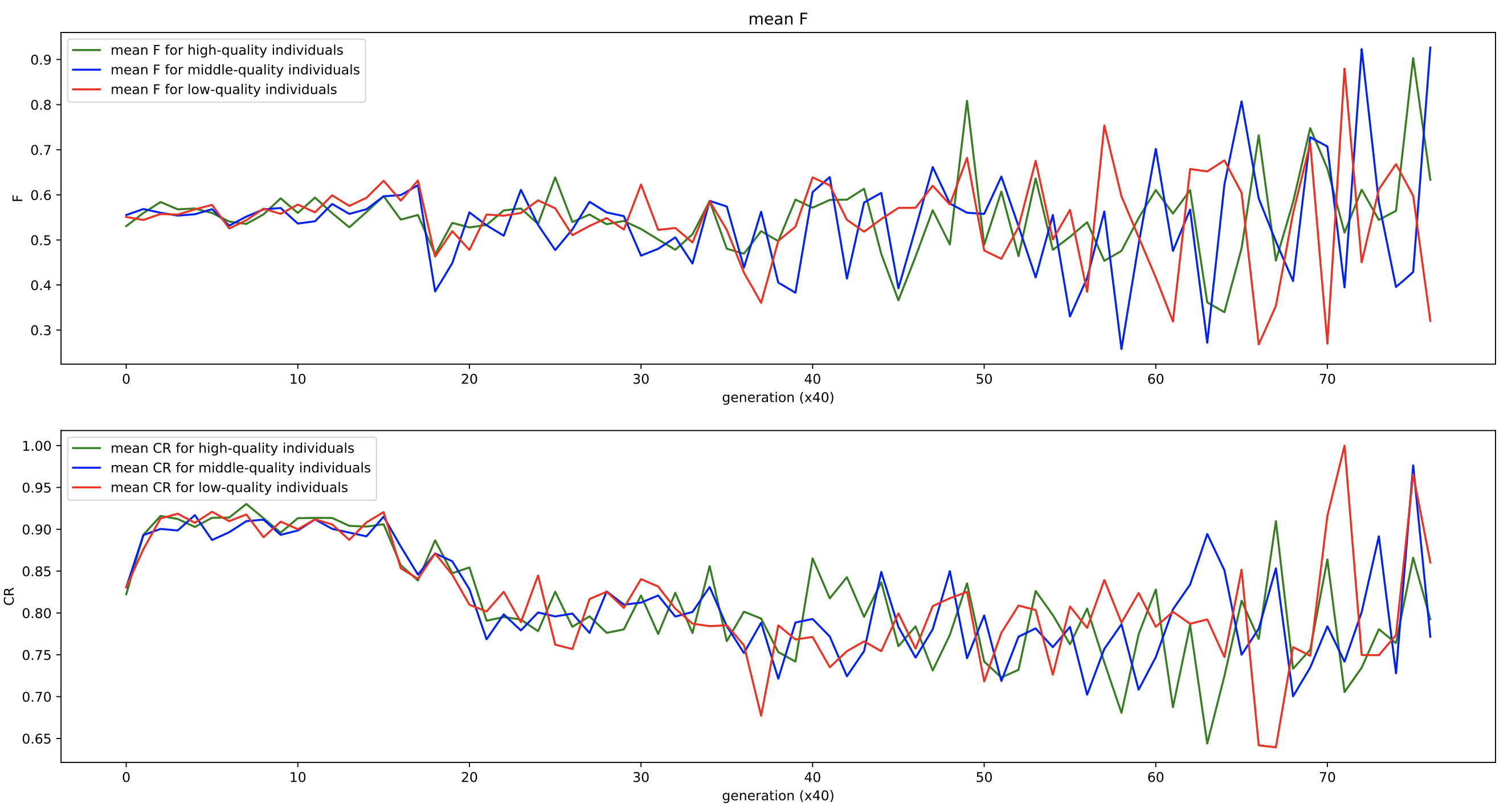}}
\caption{The evolution of $F$ and $CR$ along optimization for $f_{21}$ obtained by the learned controller. The upper (resp. lower) plot shows the $F$ (resp. $CR$) values. The population's fitness is grouped into three clusters at each generation. The associated $F$ and $CR$ values are averaged.} \label{fcrevo}
\end{figure}

Fig.~\ref{fcrevo} shows the evolution of the control parameters during the optimization obtained by the learned controller and jSO when optimizing $f_{21}$. In the figure, we show the mean values of $F$ and $CR$ at each generation by clustering ${\cal F}^t$ into three groups. The upper plot shows the mean $F$ values, while the lower plot shows the mean $CR$ values associated with the individuals in the groups. From the upper plot of Fig.~\ref{fcrevo}(a) for the learned controller, it is seen that high-quality individuals generally have a smaller $F$ value than the low-quality individuals, while the middle-quality individuals have higher $CR$ values. From Fig.~\ref{fcrevo}(b) for jSO, it is seen that along evolution, the $F$ and $CR$ values become scattered.

The better performance of LDE when running more generations indicates that the learned controller is promising for adaptive parameter control. 

\subsection{The Comparison Results on the CEC'17 Test Suite}

In this section, all 28 test problems in the CEC'13 test suite are used as the training functions. LDE is then compared with the other algorithms on the 29 functions of the CEC'17 test suite. Table~\ref{17_10D_MAXFE} and~\ref{17_30D_MAXFE} summarize the means and standard deviations of the function error values obtained by all the compared methods over 51 times on the CEC'17 test suite for 10-D and 30-D, respectively.

For 10-D test functions, LDE exhibits superiority over HSES and most of the conventional and classical DE-based algorithms, except for CoBiDE and cDE-PCM. It performs similarly to jSO. 

\begin{table*}[htbp]
	\caption{Means and standard deviations of the error values obtained by LDE and the compared algorithms on the CEC'17 benchmark suite for $n=10$ when MAXNFE has been reached or function error is less than $10^{-8}$. }
	\label{17_10D_MAXFE}
	\begin{center}
		\setlength{\tabcolsep}{1.2mm}{
			\begin{tabular}{c|cc|ccc|ccc|ccc|ccc}
				\hline
				\multirow{2}*{} &\multicolumn{2}{c|}{LDE} &\multicolumn{2}{c}{DE} & &\multicolumn{2}{c}{JADE w/o archive} & &\multicolumn{2}{c}{JADE with archive} & &\multicolumn{2}{c}{jSO} & \\ \cline{2-15}
				&Mean & Std. Dev. &Mean & Std. Dev. & WR & Mean & Std. Dev. & WR & Mean & Std. Dev. & WR & Mean & Std. Dev. & WR \\
				\hline
				$F_{1}$ &\textbf{0.00E+00} &\textbf{0.00E+00} &\textbf{0.00E+00} &\textbf{0.00E+00} &$\approx$ &\textbf{0.00E+00} &\textbf{0.00E+00} &$\approx$ &\textbf{0.00E+00} &\textbf{0.00E+00} &$\approx$ &\textbf{0.00E+00} &\textbf{0.00E+00} &$\approx$ \\
				$F_{2}$ &\textbf{0.00E+00} &\textbf{0.00E+00} &\textbf{0.00E+00} &\textbf{0.00E+00} &$\approx$ &\textbf{0.00E+00} &\textbf{0.00E+00} &$\approx$ &\textbf{0.00E+00} &\textbf{0.00E+00} &$\approx$ &\textbf{0.00E+00} &\textbf{0.00E+00} &$\approx$ \\
				$F_{3}$ &\textbf{0.00E+00} &\textbf{0.00E+00} &1.37E+00 &5.06E-01 &$-$ &\textbf{0.00E+00} &\textbf{0.00E+00} &$\approx$ &\textbf{0.00E+00} &\textbf{0.00E+00} &$\approx$ &\textbf{0.00E+00} &\textbf{0.00E+00} &$\approx$ \\
				$F_{4}$ &4.53E+00 &1.91E+00 &1.68E+01 &5.92E+00 &$-$ &3.66E+00 &1.16E+00 &$\approx$ &3.68E+00 &1.19E+00 &$\approx$ &2.07E+00 &8.08E-01 &$+$ \\
				$F_{5}$ &\textbf{0.00E+00} &\textbf{0.00E+00} &\textbf{0.00E+00} &\textbf{0.00E+00} &$\approx$ &\textbf{0.00E+00} &\textbf{0.00E+00} &$\approx$ &\textbf{0.00E+00} &\textbf{0.00E+00} &$\approx$ &\textbf{0.00E+00} &\textbf{0.00E+00} &$\approx$ \\
				$F_{6}$ &1.49E+01 &2.13E+00 &3.07E+01 &5.01E+00 &$-$ &1.34E+01 &1.38E+00 &$+$ &1.38E+01 &1.82E+00 &$+$ &1.20E+01 &5.65E-01 &$+$ \\
				$F_{7}$ &4.58E+00 &1.98E+00 &1.73E+01 &6.21E+00 &$-$ &3.85E+00 &1.00E+00 &$\approx$ &3.79E+00 &1.41E+00 &$\approx$ &2.04E+00 &7.06E-01 &$+$ \\
				$F_{8}$ &\textbf{0.00E+00} &\textbf{0.00E+00} &\textbf{0.00E+00} &\textbf{0.00E+00} &$\approx$ &1.76E-03 &1.24E-02 &$\approx$ &\textbf{0.00E+00} &\textbf{0.00E+00} &$\approx$ &\textbf{0.00E+00} &\textbf{0.00E+00} &$\approx$  \\
				$F_{9}$ &1.87E+02 &1.31E+02 &7.55E+02 &2.30E+02 &$-$ &1.26E+02 &1.06E+02 &$+$ &1.21E+02 &9.55E+01 &$+$ &\textbf{4.86E+01} &\textbf{6.38E+01} &$+$ \\
				$F_{10}$ &9.75E-02 &3.55E-01 &3.23E-01 &4.69E-01 &$-$ &1.94E+00 &9.13E-01 &$-$ &1.78E+00 &9.85E-01 &$-$ &\textbf{0.00E+00} &\textbf{0.00E+00} &$\approx$ \\
				$F_{11}$ &2.20E+01 &4.59E+01 &9.88E+00 &3.20E+01 &$\approx$ &3.18E+02 &1.74E+02 &$-$ &3.66E+02 &2.10E+02 &$-$ &2.72E+00 &1.66E+01 &$\approx$ \\
				$F_{12}$ &1.95E+00 &2.21E+00 &3.15E+00 &2.35E+00 &$-$ &5.39E+00 &3.57E+00 &$-$ &4.94E+00 &3.10E+00 &$-$ &2.05E+00 &2.31E+00 &$\approx$ \\
				$F_{13}$ &3.12E-01 &5.39E-01 &3.71E-01 &6.22E-01 &$\approx$ &2.03E+00 &5.46E+00 &$-$ &2.76E+00 &6.51E+00 &$-$ &1.56E-01 &3.62E-01 &$\approx$ \\
				$F_{14}$ &3.30E-02 &8.62E-02 &1.24E-01 &1.88E-01 &$\approx$ &2.91E-01 &1.58E-01 &$-$ &2.85E-01 &2.34E-01 &$-$ &3.14E-01 &2.08E-01 &$-$ \\
				$F_{15}$ &3.14E-01 &2.36E-01 &4.86E-01 &2.86E-01 &$-$ &7.71E+00 &2.79E+01 &$-$ &2.97E+00 &1.65E+01 &$-$ &6.46E-01 &1.98E-01 &$-$ \\
				$F_{16}$ &3.68E-01 &4.08E-01 &3.53E-01 &3.04E-01 &$\approx$ &5.00E-01 &2.84E+00 &$-$ &5.25E-01 &2.77E+00 &$-$ &5.93E-01 &4.22E-01 &$-$ \\
				$F_{17}$ &5.68E-02 &9.09E-02 &1.14E-01 &2.07E-01 &$\approx$ &5.88E+00 &8.87E+00 &$-$ &7.85E+00 &9.63E+00 &$-$ &2.43E-01 &2.09E-01 &$-$ \\
				$F_{18}$ &1.01E-02 &1.17E-02 &\textbf{4.97E-03} &\textbf{9.34E-03} &$+$ &3.49E-02 &2.07E-01 &$\approx$ &1.44E-01 &4.26E-01 &$\approx$ &3.35E-03 &6.92E-03 &$\approx$ \\
				$F_{19}$ &1.22E-02 &6.06E-02 &2.40E-01 &2.49E-01 &$-$ &4.28E-02 &1.24E-01 &$\approx$ &4.90E-01 &2.76E+00 &$-$ &3.68E-01 &1.88E-01 &$-$ \\
				$F_{20}$ &\textbf{1.25E+02} &\textbf{4.55E+01} &1.70E+02 &5.90E+01 &$-$ &1.79E+02 &4.58E+01 &$-$ &1.90E+02 &3.80E+01 &$-$ &1.42E+02 &5.05E+01 &$\approx$ \\
				$F_{21}$ &9.06E+01 &2.88E+01 &1.00E+02 &2.27E-01 &$-$ &9.62E+01 &1.94E+01 &$-$ &1.00E+02 &1.57E-01 &$-$ &9.88E+01 &8.54E+00 &$\approx$ \\
				$F_{22}$ &3.06E+02 &3.04E+00 &3.04E+02 &2.78E+00 &$+$ &3.06E+02 &1.98E+00 &$\approx$ &3.05E+02 &1.69E+00 &$\approx$ &\textbf{3.01E+02} &\textbf{1.24E+00} &$+$ \\
				$F_{23}$ &\textbf{2.42E+02} &\textbf{1.16E+02} &3.06E+02 &7.83E+01 &$-$ &3.19E+02 &5.47E+01 &$\approx$ &3.11E+02 &6.96E+01 &$-$ &2.70E+02 &1.00E+02 &$-$ \\
				$F_{24}$ &4.07E+02 &1.80E+01 &4.17E+02 &2.30E+01 &$-$ &4.28E+02 &2.18E+01 &$-$ &4.22E+02 &2.71E+01 &$-$ &4.08E+02 &1.87E+01 &$\approx$ \\
				$F_{25}$ &\textbf{3.00E+02} &\textbf{0.00E+00} &\textbf{3.00E+02} &\textbf{0.00E+00} &$\approx$ &3.00E+02 &2.55E+01 &$\approx$ &3.38E+02 &1.76E+02 &$\approx$ &\textbf{3.00E+02} &\textbf{0.00E+00} &$\approx$ \\
				$F_{26}$ &3.90E+02 &1.38E+00 &3.92E+02 &2.61E+00 &$-$ &3.92E+02 &2.86E+00 &$-$ &3.91E+02 &6.93E+00 &$\approx$ &\textbf{3.89E+02} &\textbf{3.56E-01} &$+$ \\
				$F_{27}$ &3.00E+02 &5.78E+01 &3.51E+02 &1.09E+02 &$\approx$ &4.19E+02 &1.40E+02 &$-$ &4.74E+02 &1.49E+02 &$-$ &\textbf{3.00E+02} &\textbf{0.00E+00} &$\approx$ \\
				$F_{28}$ &2.33E+02 &4.39E+00 &2.36E+02 &5.75E+00 &$-$ &2.56E+02 &2.06E+01 &$-$ &2.48E+02 &9.30E+00 &$-$ &2.37E+02 &2.93E+00 &$-$ \\
				$F_{29}$ &3.97E+02 &8.13E+00 &4.54E+02 &8.53E+01 &$-$ &1.45E+05 &3.11E+05 &$-$ &1.58E+05 &3.34E+05 &$-$ &\textbf{3.93E+02} &\textbf{2.87E+00} &$+$ \\
				\hline
				{} &\multicolumn{2}{c}{$+/\approx/-$} &\multicolumn{3}{|c|}{2/11/16} &\multicolumn{3}{|c|}{2/12/15} &\multicolumn{3}{|c|}{2/11/16} &\multicolumn{3}{c}{7/15/7} \\
				\hline
		\end{tabular}}
		
		\vspace{2em}
		\setlength{\tabcolsep}{0.97mm}{
			\begin{tabular}{c|ccc|ccc|ccc|ccc|ccc}
				\hline
				\multirow{2}*{} &\multicolumn{2}{c}{CoBiDE} & &\multicolumn{2}{c}{CoBiDE-PCM} & &\multicolumn{2}{c}{cDE} & &\multicolumn{2}{c}{cDE-PCM} & &\multicolumn{2}{c}{HSES} & \\ 	\cline{2-16}		
				& Mean & Std. Dev. & WR & Mean & Std. Dev. & WR & Mean & Std. Dev. & WR & Mean & Std. Dev. & WR & Mean & Std. Dev. & WR \\
				\hline
				$F_{1}$ &\textbf{0.00E+00} &\textbf{0.00E+00} &$\approx$ &\textbf{0.00E+00} &\textbf{0.00E+00} &$\approx$ &\textbf{0.00E+00} &\textbf{0.00E+00} &$\approx$ &\textbf{0.00E+00} &\textbf{0.00E+00} &$\approx$ &\textbf{0.00E+00} &\textbf{0.00E+00} &$\approx$ \\
				$F_{2}$ &\textbf{0.00E+00} &\textbf{0.00E+00} &$\approx$ &\textbf{0.00E+00} &\textbf{0.00E+00} &$\approx$ &\textbf{0.00E+00} &\textbf{0.00E+00} &$\approx$ &\textbf{0.00E+00} &\textbf{0.00E+00} &$\approx$ &\textbf{0.00E+00} &\textbf{0.00E+00} &$\approx$ \\
				$F_{3}$ &\textbf{0.00E+00} &\textbf{0.00E+00} &$\approx$ &\textbf{0.00E+00} &\textbf{0.00E+00} &$\approx$ &\textbf{0.00E+00} &\textbf{0.00E+00} &$\approx$ &\textbf{0.00E+00} &\textbf{0.00E+00} &$\approx$ &\textbf{0.00E+00} &\textbf{0.00E+00} &$\approx$ \\
				$F_{4}$ &3.57E+00 &1.52E+00 &$+$ &1.14E+01 &5.01E+00 &$-$ &4.63E+00 &1.83E+00 &$\approx$ &3.47E+00 &1.10E+00 &$+$ &\textbf{1.01E+00} &\textbf{8.92E-01} &$+$\\
				$F_{5}$  &\textbf{0.00E+00} &\textbf{0.00E+00} &$\approx$ &2.68E-06 &5.33E-06 &$-$ &\textbf{0.00E+00} &\textbf{0.00E+00} &$\approx$ &\textbf{0.00E+00} &\textbf{0.00E+00} &$\approx$ &\textbf{0.00E+00} &\textbf{0.00E+00} &$\approx$ \\
				$F_{6}$ &1.35E+01 &1.99E+00 &$+$ &1.94E+01 &5.70E+00 &$-$ &1.55E+01 &2.71E+00 &$\approx$ &1.45E+01 &1.48E+00 &$\approx$ &\textbf{1.14E+01} &\textbf{7.20E-01} &$+$\\
				$F_{7}$ &3.77E+00 &1.70E+00 &$+$ &1.29E+01 &5.72E+00 &$-$ &5.74E+00 &2.09E+00 &$-$ &3.62E+00 &1.62E+00 &$+$ &\textbf{5.46E-01} &\textbf{7.71E-01} &$+$\\
				$F_{8}$ &\textbf{0.00E+00} &\textbf{0.00E+00} &$\approx$ &7.65E-02 &1.69E-01 &$\approx$ &\textbf{0.00E+00} &\textbf{0.00E+00} &$\approx$ &\textbf{0.00E+00} &\textbf{0.00E+00} &$\approx$ &\textbf{0.00E+00} &\textbf{0.00E+00} &$\approx$ \\
				$F_{9}$ &9.18E+01 &9.47E+01 &$+$ &4.01E+02 &2.37E+02 &$-$ &1.88E+02 &1.09E+02 &$\approx$ &1.80E+02 &7.37E+01 &$\approx$ &1.04E+02 &1.56E+02 &$+$\\
				$F_{10}$ &9.75E-02 &2.96E-01 &$\approx$ &5.18E+00 &5.39E+00 &$-$ &5.55E-01 &9.06E-01 &$-$  &\textbf{0.00E+00} &\textbf{0.00E+00} &$\approx$ &7.80E-02 &2.67E-01 &$\approx$\\
				$F_{11}$ &\textbf{1.67E-01} &\textbf{1.30E-01} &$+$ &3.62E+02 &2.01E+02 &$-$ &1.39E+02 &1.59E+02 &$-$ &5.83E+00 &2.31E+01 &$\approx$ &1.09E+01 &3.10E+01 &$\approx$\\
				$F_{12}$  &\textbf{1.07E+00} &\textbf{1.82E+00} &$+$ &8.38E+00 &5.47E+00 &$-$ &5.05E+00 &2.95E+00 &$-$ &1.63E+00 &2.13E+00 &$\approx$ &3.44E+00 &2.48E+00 &$-$\\
				$F_{13}$  &\textbf{0.00E+00} &\textbf{0.00E+00} &$+$ &9.58E+00 &1.02E+01 &$-$ &6.49E-01 &9.64E-01 &$-$ &1.95E-02 &1.38E-01 &$+$ &6.92E+00 &3.24E+01 &$-$\\
				$F_{14}$  &\textbf{9.31E-03} &\textbf{3.49E-02} &$+$ &2.57E+00 &2.69E+00 &$-$ &3.09E-01 &4.80E-01 &$-$ &6.38E-03 &9.38E-03 &$\approx$ &5.59E-01 &7.50E-01 &$-$\\
				$F_{15}$  &\textbf{2.25E-01} &\textbf{1.51E-01} &$\approx$ &3.65E+01 &5.42E+01 &$-$ &2.96E+00 &1.67E+01 &$\approx$ &2.99E-01 &1.29E-01 &$\approx$ &3.07E+00 &1.64E+01 &$-$\\
				$F_{16}$  &1.57E+00 &6.12E-01 &$-$ &1.32E+01 &2.39E+01 &$-$ &1.33E+00 &8.04E+00 &$-$ &\textbf{8.27E-02} &\textbf{1.04E-01} &$+$ &1.66E+01 &1.09E+01 &$-$\\
				$F_{17}$  &\textbf{6.38E-03} &\textbf{2.75E-02} &$+$ &2.24E+01 &2.45E+01 &$-$ &1.14E+00 &3.89E+00 &$-$ &7.34E-02 &1.45E-01 &$\approx$ &5.24E-01 &4.11E-01 &$-$\\
				$F_{18}$  &8.99E-03 &1.33E-02 &$\approx$ &1.12E+00 &9.42E-01 &$-$ &4.20E-03 &8.00E-03 &$\approx$ &1.59E-02 &1.24E-02 &$-$ &8.22E-01 &1.75E+00 &$-$\\
				$F_{19}$  &\textbf{0.00E+00} &\textbf{0.00E+00} &$\approx$ &6.25E+00 &2.35E+01 &$-$ &9.91E-02 &1.96E-01 &$\approx$ &\textbf{0.00E+00} &\textbf{0.00E+00} &$\approx$ &1.24E+01 &1.06E+01 &$-$\\
				$F_{20}$  &1.38E+02 &5.10E+01 &$\approx$ &1.91E+02 &4.52E+01 &$-$ &1.79E+02 &4.86E+01 &$-$ &1.44E+02 &5.22E+01 &$\approx$ &1.91E+02 &3.04E+01 &$-$\\
				$F_{21}$  &\textbf{7.86E+01} &\textbf{4.12E+01} &$\approx$ &9.70E+01 &2.02E+01 &$-$ &9.28E+01 &2.63E+01 &$-$ &9.22E+01 &2.69E+01 &$\approx$ &1.00E+02 &0.00E+00 &$\approx$\\
				$F_{22}$ &3.05E+02 &1.57E+00 &$\approx$ &3.12E+02 &4.90E+00 &$-$ &3.07E+02 &2.54E+00 &$-$ &3.05E+02 &2.27E+00 &$\approx$ &3.01E+02 &1.67E+00 &$+$\\
				$F_{23}$  &2.74E+02 &1.02E+02 &$\approx$ &3.25E+02 &6.59E+01 &$-$ &3.26E+02 &4.58E+01 &$-$ &2.79E+02 &9.96E+01 &$\approx$ &3.28E+02 &7.42E-01 &$\approx$\\
				$F_{24}$ &4.00E+02 &9.06E+00 &$\approx$ &4.20E+02 &2.31E+01 &$-$ &4.17E+02 &2.32E+01 &$-$ &\textbf{4.01E+02} &\textbf{1.07E+01} &$+$ &4.46E+02 &1.02E+00 &$-$\\
				$F_{25}$  &\textbf{3.00E+02} &\textbf{0.00E+00} &$\approx$ &3.30E+02 &1.36E+02 &$\approx$ &3.19E+02 &1.29E+02 &$\approx$ &\textbf{3.00E+02} &\textbf{0.00E+00} &$\approx$ &\textbf{3.00E+02} &\textbf{0.00E+00} &$\approx$\\
				$F_{26}$ &3.89E+02 &8.94E-01 &$+$ &3.98E+02 &1.68E+01 &$-$ &3.90E+02 &1.99E+00 &$\approx$ &3.89E+02 &9.90E-01 &$+$ &3.97E+02 &1.81E+00 &$-$\\
				$F_{27}$  &3.06E+02 &3.93E+01 &$\approx$ &4.32E+02 &1.37E+02 &$-$ &4.04E+02 &1.38E+02 &$-$ &3.11E+02 &5.51E+01 &$\approx$ &5.99E+02 &2.55E+01 &$-$\\
				$F_{28}$  &\textbf{2.30E+02} &\textbf{2.75E+00} &$+$ &2.53E+02 &2.37E+01 &$-$ &2.36E+02 &6.37E+00 &$-$ &2.32E+02 &4.27E+00 &$\approx$ &2.64E+02 &1.09E+01 &$-$\\
				$F_{29}$  &3.96E+02 &3.47E+00 &$+$ &2.50E+05 &4.16E+05 &$-$ &8.06E+04 &2.43E+05 &$-$ &4.04E+02 &1.80E+01 &$-$ &4.17E+02 &2.49E+01 &$-$\\
				\hline
				{} &\multicolumn{3}{|c|}{12/16/1} &\multicolumn{3}{|c|}{0/5/24} &\multicolumn{3}{|c|}{0/13/16} &\multicolumn{3}{|c|}{6/21/2}  &\multicolumn{3}{c}{5/10/14}\\
				\hline
		\end{tabular}}
	\end{center}
\end{table*}

\begin{table*}
	\caption{Means and standard deviations of the error values obtained by LDE and the compared algorithms on the CEC'17 benchmark suite for $n=30$ when MAXNFE has been reached or function error is less than $10^{-8}$. }
	\label{17_30D_MAXFE}
	\begin{center}
		\setlength{\tabcolsep}{1.2mm}{
			\begin{tabular}{c|cc|ccc|ccc|ccc|ccc}
				\hline
				\multirow{2}*{} &\multicolumn{2}{c|}{LDE} &\multicolumn{2}{c}{DE} & &\multicolumn{2}{c}{JADE w/o archive} & &\multicolumn{2}{c}{JADE with archive} & &\multicolumn{2}{c}{jSO} & \\ \cline{2-15}
				&Mean & Std. Dev. &Mean & Std. Dev. & WR & Mean & Std. Dev. & WR & Mean & Std. Dev. & WR & Mean & Std. Dev. & WR\\
				\hline
				$F_{1}$ &\textbf{0.00E+00} &\textbf{0.00E+00} &9.21E+00 &3.42E+00 &$-$ &\textbf{0.00E+00} &\textbf{0.00E+00} &$\approx$ &\textbf{0.00E+00} &\textbf{0.00E+00} &$\approx$ &\textbf{0.00E+00} &\textbf{0.00E+00} &$\approx$ \\
				$F_{2}$ &3.25E-05 &6.70E-05 &4.20E+04 &6.20E+03 &$-$ &6.02E+03 &1.34E+04 &$-$ &6.96E+03 &1.51E+04 &$-$ &\textbf{0.00E+00} &\textbf{0.00E+00} &$+$ \\
				$F_{3}$ &4.41E+01 &2.65E+01 &6.11E+01 &5.07E+00 &$-$ &2.98E+01 &2.97E+01 &$+$ &4.26E+01 &2.70E+01 &$+$ &5.86E+01 &0.00E+00 &$\approx$ \\
				$F_{4}$ &3.71E+01 &7.35E+00 &1.83E+02 &7.53E+00 &$-$ &2.71E+01 &4.34E+00 &$+$ &2.67E+01 &4.26E+00 &$+$ &9.51E+00 &1.93E+00 &$+$ \\
				$F_{5}$ &1.66E-04 &6.07E-05 &8.86E-04 &1.98E-04 &$-$ &\textbf{0.00E+00} &\textbf{0.00E+00} &$+$ &\textbf{0.00E+00} &\textbf{0.00E+00} &$+$ &8.24E-09 &3.29E-08 &$+$ \\
				$F_{6}$ &6.66E+01 &8.80E+00 &2.26E+02 &8.94E+00 &$-$ &5.50E+01 &3.79E+00 &$+$ &5.37E+01 &3.60E+00 &$+$ &\textbf{3.89E+01} &\textbf{1.72E+00} &$+$ \\
				$F_{7}$ &3.86E+01 &9.40E+00 &1.86E+02 &9.06E+00 &$-$ &2.64E+01 &4.17E+00 &$+$ &2.50E+01 &4.41E+00 &$+$ &9.44E+00 &1.81E+00 &$+$ \\
				$F_{8}$ &\textbf{0.00E+00} &\textbf{0.00E+00} &2.35E-09 &7.56E-09 &$\approx$ &5.27E-03 &2.11E-02 &$\approx$ &3.37E-02 &1.08E-01 &$\approx$ &\textbf{0.00E+00} &\textbf{0.00E+00} &$\approx$ \\
				$F_{9}$ &1.80E+03 &3.63E+02 &6.81E+03 &2.76E+02 &$-$ &1.89E+03 &1.93E+02 &$\approx$ &1.87E+03 &2.55E+02 &$\approx$ &1.58E+03 &2.25E+02 &$+$ \\
				$F_{10}$ &1.20E+01 &4.33E+00 &7.57E+01 &1.86E+01 &$-$ &3.10E+01 &2.67E+01 &$-$ &3.12E+01 &2.50E+01 &$-$ &\textbf{3.63E+00} &\textbf{2.12E+00} &$+$ \\
				$F_{11}$ &2.13E+03 &2.05E+03 &9.85E+05 &3.93E+05 &$-$ &2.92E+03 &2.95E+03 &$-$ &1.21E+03 &4.25E+02 &$\approx$ &1.34E+02 &1.06E+02 &$+$ \\
				$F_{12}$ &2.28E+01 &8.46E+00 &1.93E+02 &2.16E+01 &$-$ &4.27E+01 &2.96E+01 &$-$ &1.52E+03 &6.29E+03 &$-$ &\textbf{1.24E+01} &\textbf{8.22E+00} &$+$ \\
				$F_{13}$ &2.64E+01 &9.72E+00 &9.53E+01 &7.13E+00 &$-$ &2.32E+03 &1.03E+04 &$-$ &5.78E+03 &9.17E+03 &$-$ &2.29E+01 &1.55E+00 &$\approx$ \\
				$F_{14}$ &7.45E+00 &2.40E+00 &6.22E+01 &5.56E+00 &$-$ &3.99E+02 &2.64E+03 &$-$ &1.28E+03 &4.06E+03 &$-$ &\textbf{2.29E+00} &\textbf{1.24E+00} &$+$ \\
				$F_{15}$ &4.18E+02 &1.72E+02 &1.13E+03 &1.35E+02 &$-$ &4.07E+02 &1.44E+02 &$\approx$ &4.12E+02 &1.44E+02 &$\approx$ &\textbf{1.21E+02} &\textbf{1.04E+02} &$+$ \\
				$F_{16}$ &4.10E+01 &1.94E+01 &3.44E+02 &5.56E+01 &$-$ &6.82E+01 &1.41E+01 &$-$ &7.18E+01 &2.80E+01 &$-$ &\textbf{3.46E+01} &\textbf{5.89E+00} &$\approx$ \\
				$F_{17}$ &2.31E+01 &4.07E+00 &1.49E+02 &2.27E+01 &$-$ &1.51E+04 &4.34E+04 &$-$ &1.16E+04 &3.16E+04 &$-$ &2.12E+01 &5.39E-01 &$+$ \\
				$F_{18}$ &5.99E+00 &9.65E-01 &2.38E+01 &2.57E+00 &$-$ &1.15E+01 &5.32E+00 &$-$ &1.70E+03 &4.61E+03 &$-$ &\textbf{3.39E+00} &\textbf{6.35E-01} &$+$ \\
				$F_{19}$ &5.62E+01 &5.50E+01 &1.59E+02 &8.72E+01 &$-$ &1.07E+02 &5.37E+01 &$-$ &1.11E+02 &5.41E+01 &$-$ &3.17E+01 &6.69E+00 &$\approx$ \\
				$F_{20}$ &2.39E+02 &8.39E+00 &3.75E+02 &1.01E+01 &$-$ &2.28E+02 &4.31E+00 &$+$ &2.26E+02 &4.55E+00 &$+$ &\textbf{2.09E+02} &\textbf{2.09E+00} &$+$ \\
				$F_{21}$ &\textbf{1.00E+02} &\textbf{0.00E+00} &1.00E+02 &1.56E-07 &$-$ &\textbf{1.00E+02} &\textbf{0.00E+00} &$\approx$ &1.00E+02 &2.33E+00 &$\approx$ &\textbf{1.00E+02} &\textbf{0.00E+00} &$\approx$ \\
				$F_{22}$ &3.85E+02 &1.00E+01 &5.26E+02 &1.04E+01 &$-$ &3.72E+02 &5.38E+00 &$+$ &3.73E+02 &5.77E+00 &$+$ &\textbf{3.48E+02} &\textbf{3.57E+00} &$+$ \\
				$F_{23}$ &4.65E+02 &1.19E+01 &5.95E+02 &7.42E+00 &$-$ &4.39E+02 &4.61E+00 &$+$ &4.42E+02 &5.67E+00 &$+$ &4.24E+02 &2.02E+00 &$+$ \\
				$F_{24}$ &3.87E+02 &6.48E-01 &3.87E+02 &6.63E-02 &$-$ &3.87E+02 &2.12E-01 &$-$ &3.87E+02 &1.52E-01 &$-$ &3.87E+02 &1.42E-02 &$-$ \\
				$F_{25}$ &1.25E+03 &4.05E+02 &2.72E+03 &1.02E+02 &$-$ &1.17E+03 &1.38E+02 &$+$ &1.19E+03 &6.15E+01 &$+$ &8.95E+02 &2.91E+01 &$+$ \\
				$F_{26}$ &4.97E+02 &8.45E+00 &\textbf{4.84E+02} &\textbf{1.28E+01} &$+$ &5.03E+02 &7.62E+00 &$-$ &5.04E+02 &6.44E+00 &$-$ &4.92E+02 &8.09E+00 &$+$ \\
				$F_{27}$ &3.06E+02 &2.52E+01 &3.30E+02 &3.87E+01 &$-$ &3.37E+02 &5.38E+01 &$-$ &3.40E+02 &5.74E+01 &$\approx$ &\textbf{3.00E+02} &\textbf{0.00E+00} &$+$ \\
				$F_{28}$ &4.37E+02 &3.42E+01 &9.58E+02 &9.09E+01 &$-$ &4.80E+02 &3.21E+01 &$-$ &4.77E+02 &2.73E+01 &$-$ &4.37E+02 &1.76E+01 &$\approx$ \\
				$F_{29}$ &1.99E+03 &4.46E+01 &8.32E+03 &1.55E+03 &$-$ &2.18E+03 &1.59E+02 &$-$ &2.13E+03 &1.62E+02 &$-$ &\textbf{1.96E+03} &\textbf{1.10E+01} &$+$ \\
				\hline
				{} &\multicolumn{2}{c}{$+/\approx/-$} &\multicolumn{3}{|c|}{1/1/27} &\multicolumn{3}{|c|}{9/5/15} &\multicolumn{3}{|c|}{9/7/13} &\multicolumn{3}{c}{20/8/1} \\
				\hline
		\end{tabular}}

		\vspace{2em}
		\setlength{\tabcolsep}{0.97mm}{
			\begin{tabular}{c|ccc|ccc|ccc|ccc|ccc}
				\hline
				\multirow{2}*{} &\multicolumn{2}{c}{CoBiDE} & &\multicolumn{2}{c}{CoBiDE-PCM} & &\multicolumn{2}{c}{cDE} & &\multicolumn{2}{c}{cDE-PCM} & &\multicolumn{2}{c}{HSES} & \\ \cline{2-16}
				& Mean & Std. Dev. & WR & Mean & Std. Dev. & WR & Mean & Std. Dev. & WR & Mean & Std. Dev. & WR & Mean & Std. Dev. & WR \\
				\hline
				$F_{1}$  &\textbf{0.00E+00} &\textbf{0.00E+00} &$\approx$ &\textbf{0.00E+00} &\textbf{0.00E+00} &$\approx$ &\textbf{0.00E+00} &\textbf{0.00E+00} &$\approx$ &\textbf{0.00E+00} &\textbf{0.00E+00} &$\approx$ &\textbf{0.00E+00} &\textbf{0.00E+00} &$\approx$ \\
				$F_{2}$  &\textbf{0.00E+00} &\textbf{0.00E+00} &$+$ &\textbf{0.00E+00} &\textbf{0.00E+00} &$+$ &\textbf{0.00E+00} &\textbf{0.00E+00} &$+$ &\textbf{0.00E+00} &\textbf{0.00E+00} &$+$ &\textbf{0.00E+00} &\textbf{0.00E+00} &$+$\\
				$F_{3}$  &4.20E+01 &2.79E+01 &$\approx$ &2.83E+01 &2.94E+01 &$+$ &5.69E+01 &1.08E+01 &$\approx$ &3.23E+01 &2.98E+01 &$+$ &\textbf{4.31E+00} &\textbf{8.62E+00} &$+$ \\
				$F_{4}$ &3.99E+01 &9.66E+00 &$\approx$ &7.80E+01 &2.12E+01 &$-$ &5.07E+01 &6.12E+00 &$-$ &4.86E+01 &8.45E+00 &$-$ &\textbf{8.76E+00} &\textbf{2.65E+00} &$+$\\
				$F_{5}$  &3.49E-08 &4.73E-08 &$+$ &2.57E-02 &6.51E-02 &$-$ &\textbf{0.00E+00} &\textbf{0.00E+00} &$+$ &3.73E-02 &1.60E-01 &$\approx$ &\textbf{0.00E+00} &\textbf{0.00E+00} &$+$\\
				$F_{6}$ &7.12E+01 &1.00E+01 &$-$ &1.15E+02 &2.44E+01 &$-$ &9.21E+01 &6.39E+00 &$-$ &8.74E+01 &9.67E+00 &$-$ &4.07E+01 &3.81E+00 &$+$\\
				$F_{7}$ &3.92E+01 &1.07E+01 &$\approx$ &7.27E+01 &2.03E+01 &$-$ &5.82E+01 &9.09E+00 &$-$ &4.99E+01 &7.49E+00 &$-$ &\textbf{7.88E+00} &\textbf{2.87E+00} &$+$\\
				$F_{8}$ &0.00E+00 &0.00E+00 &$\approx$ &3.10E+01 &4.44E+01 &$-$ &7.53E-01 &1.43E+00 &$-$ &3.04E+00 &2.23E+00 &$-$ &\textbf{0.00E+00} &\textbf{0.00E+00} &$\approx$\\
				$F_{9}$ &1.82E+03 &4.52E+02 &$\approx$ &2.81E+03 &5.74E+02 &$-$ &2.33E+03 &2.84E+02 &$-$ &2.72E+03 &2.97E+02 &$-$ &\textbf{9.90E+02} &\textbf{3.86E+02} &$+$\\
				$F_{10}$  &1.62E+01 &9.72E+00 &$-$ &1.23E+02 &6.39E+01 &$-$ &2.02E+01 &1.19E+01 &$-$ &1.74E+02 &6.24E+01 &$-$ &1.37E+01 &2.14E+01 &$-$\\
				$F_{11}$  &2.83E+03 &5.16E+03 &$\approx$ &1.18E+04 &1.08E+04 &$-$ &2.02E+04 &1.43E+04 &$-$ &1.11E+04 &1.11E+04 &$-$ &\textbf{4.42E+01} &\textbf{1.00E+02} &$+$\\
				$F_{12}$  &2.51E+01 &8.58E+00 &$\approx$ &1.12E+02 &1.30E+02 &$-$ &5.27E+01 &2.28E+01 &$-$ &2.28E+03 &8.37E+03 &$-$ &2.90E+01 &1.33E+01 &$\approx$\\
				$F_{13}$  &\textbf{1.09E+01} &\textbf{4.66E+00} &$+$ &1.82E+02 &7.05E+01 &$-$ &3.41E+01 &9.14E+00 &$-$ &2.21E+02 &6.77E+01 &$-$ &1.43E+01 &1.08E+01 &$+$\\
				$F_{14}$  &6.85E+00 &2.77E+00 &$\approx$ &1.68E+02 &1.19E+02 &$-$ &1.86E+01 &4.77E+00 &$-$ &3.50E+02 &1.50E+02 &$-$ &5.59E+00 &3.71E+00 &$+$\\
				$F_{15}$  &3.78E+02 &1.46E+02 &$\approx$ &7.32E+02 &2.57E+02 &$-$ &4.88E+02 &1.35E+02 &$-$ &4.56E+02 &2.13E+02 &$\approx$ &2.33E+02 &2.04E+02 &$+$\\
				$F_{16}$ &4.38E+01 &3.27E+01 &$\approx$ &2.50E+02 &1.54E+02 &$-$ &9.68E+01 &2.88E+01 &$-$ &1.61E+02 &1.24E+02 &$-$ &5.93E+01 &1.06E+02 &$-$\\
				$F_{17}$  &\textbf{1.86E+01} &\textbf{8.57E+00} &$+$ &9.05E+01 &6.82E+01 &$-$ &3.16E+01 &5.41E+00 &$-$ &8.72E+02 &1.29E+03 &$-$ &2.07E+01 &5.78E+00 &$+$\\
				$F_{18}$  &4.78E+00 &1.43E+00 &$+$ &1.14E+02 &6.85E+01 &$-$ &1.54E+01 &2.29E+00 &$-$ &2.21E+02 &8.50E+01 &$-$ &3.88E+00 &1.59E+00 &$+$\\
				$F_{19}$  &\textbf{4.31E+01} &\textbf{5.74E+01} &$+$ &2.37E+02 &1.27E+02 &$-$ &1.00E+02 &5.43E+01 &$-$ &1.11E+02 &8.17E+01 &$-$ &1.57E+02 &5.04E+01 &$-$\\
				$F_{20}$  &2.43E+02 &9.18E+00 &$-$ &2.75E+02 &1.86E+01 &$-$ &2.58E+02 &7.12E+00 &$-$ &2.50E+02 &1.07E+01 &$-$ &2.09E+02 &3.93E+00 &$+$\\
				$F_{21}$  &\textbf{1.00E+02} &\textbf{0.00E+00} &$\approx$ &5.74E+02 &1.12E+03 &$-$ &2.08E+02 &5.32E+02 &$\approx$ &1.01E+02 &1.54E+00 &$\approx$ &\textbf{1.00E+02} &\textbf{0.00E+00} &$\approx$\\
				$F_{22}$  &3.88E+02 &8.88E+00 &$-$ &4.25E+02 &2.63E+01 &$-$ &3.98E+02 &6.50E+00 &$-$ &3.99E+02 &1.49E+01 &$-$ &3.52E+02 &8.46E+00 &$+$\\
				$F_{23}$  &4.64E+02 &1.17E+01 &$\approx$ &4.97E+02 &2.11E+01 &$-$ &4.80E+02 &8.36E+00 &$-$ &4.81E+02 &2.54E+01 &$-$ &\textbf{4.19E+02} &\textbf{5.49E+00} &$+$\\
				$F_{24}$  &3.87E+02 &4.68E-01 &$-$ &3.88E+02 &2.57E+00 &$-$ &3.87E+02 &4.70E-01 &$-$ &3.91E+02 &1.00E+01 &$-$ &\textbf{3.87E+02} &\textbf{2.67E-02} &$\approx$\\
				$F_{25}$  &1.37E+03 &2.88E+02 &$\approx$ &1.94E+03 &3.74E+02 &$-$ &1.55E+03 &2.01E+02 &$-$ &1.48E+03 &4.62E+02 &$-$ &\textbf{8.93E+02} &\textbf{1.44E+02} &$+$\\
				$F_{26}$  &4.97E+02 &1.03E+01 &$\approx$ &5.25E+02 &1.39E+01 &$-$ &4.96E+02 &1.23E+01 &$\approx$ &5.43E+02 &2.70E+01 &$-$ &5.17E+02 &8.85E+00 &$-$\\
				$F_{27}$  &3.28E+02 &4.72E+01 &$-$ &3.57E+02 &6.20E+01 &$\approx$ &3.24E+02 &4.59E+01 &$-$ &3.60E+02 &6.40E+01 &$\approx$ &3.24E+02 &4.38E+01 &$-$\\
				$F_{28}$  &\textbf{4.30E+02} &\textbf{4.64E+01} &$+$ &6.66E+02 &1.36E+02 &$-$ &5.20E+02 &4.85E+01 &$-$ &5.81E+02 &1.44E+02 &$-$ &4.65E+02 &6.57E+01 &$\approx$\\
				$F_{29}$  &2.06E+03 &8.33E+01 &$-$ &2.22E+03 &2.39E+02 &$-$ &2.13E+03 &1.44E+02 &$-$ &2.38E+03 &2.38E+02 &$-$ &2.05E+03 &3.35E+01 &$-$\\
				\hline
				{} &\multicolumn{3}{|c|}{7/15/7} &\multicolumn{3}{|c|}{2/2/25} &\multicolumn{3}{|c|}{2/4/23} &\multicolumn{3}{|c|}{2/5/22}&\multicolumn{3}{c}{17/6/6}\\
				\hline
		\end{tabular}}
	\end{center}
\end{table*}

\begin{table*}
	\caption{Average ranking of the compared algorithms according to their APS values on the CEC'17 benchmark functions}\label{ranking}
	\begin{center}
		\setlength{\tabcolsep}{1.3mm}{
			\begin{tabular}{cc|c|c|c|c|c|c|c|c|c|c}
				\hline
				\multicolumn{2}{c|}{Alg.} &jSO &CoBiDE &LDE  &HSES &JADE w/o archive &JADE with archive &cDE-PCM &cDE &DE &CoBiDE-PCM    \\
				\hline
				\multirow{3}*{Rank} &$n=10$ &1.4482 &\textbf{0.7586} &1.2414 &4.0345 &3.5172 &3.5172 &1.0690 &3.7241 &3.5517 &6.3793 \\
				&$n=30$ &\textbf{0.5517} &2.1379 &2.3793 &1.0345 &3.1034 &3.1379 &5.8276 &4.4138 &7.1379 &6.6207  \\
				&Avg. &\textbf{1.0} &1.4483 &1.8103 &2.5345 &3.3103 &3.3276 &3.4483 &4.0690 &5.3448 &6.5 \\
				\hline
		\end{tabular}}
	\end{center}
\end{table*}	

Particularly, LDE performs better than the classical DE and JADE with archive on 16 functions, CoBiDE-PCM on 24 functions, cDE on 16 functions, and HSES on 14 functions. LDE performs worse than CoBiDE on 12 functions,  jSO on 7 functions and cDE-PCM on 6 functions. LDE performs similar to CoBiDE and jSO on 16 and 15 functions, respectively.


For 30D test problems, it is seen that jSO and HSES perform better than LDE on most of the test functions. However, LDE performs better than the rest of the algorithms in general. Particularly, LDE performs better than classical DE and the other adaptive DEs on more functions than that it performs worse than these algorithms. The performance of LDE is similar to CoBiDE in the sense that the numbers of functions that LDE outperforms CoBiDE and CoBiDE outperforms LDE are the same. 

The ranking result of all algorithms is shown in Table~\ref{ranking}. It is seen that LDE ranks the third on the CEC'17 benchmark suite. Note that first the agent is learned from CEC'13. Its performance on CEC'17 implies that indeed some useful knowledge which is helpful for parameter control is effectively learned. Second, once the controller has been learned, it is applied to solve new test functions without requiring any tuning of the algorithmic parameters. This can greatly reduce possibly large amount of computational efforts.

\section{Sensitivity Analysis}\label{sen}

One of the main parameters that greatly influence the performance of LDE is the number of neurons (i.e. the sizes of ${\cal H}^t$ and ${\cal C}^t$) used in the hidden layers. A higher number can increase the representation ability of the LSTM but may cause over-fitting. Here we investigate the effect of different neuron sizes to the performance of LDE on the last eight functions $f_{21}-f_{28}$ of CEC'13 for 10-D and 30-D.

\begin{table*}[htbp]
	\caption{The results of last eight problems of CEC'13 with different cell size for $n=10$ at termination}
	\label{13_10D_CELLS_MAXFE}
	\begin{center}
		\setlength{\tabcolsep}{0.95mm}{
			\begin{tabular}{c|cc|cc|cc|cc|cc|cc}
				\hline
				cell size &\multicolumn{2}{c|}{500} &\multicolumn{2}{c|}{1000} &\multicolumn{2}{c|}{1500} &\multicolumn{2}{c|}{2000} &\multicolumn{2}{c|}{2500} &\multicolumn{2}{c}{3000} \\ \hline
				& Mean & Std. Dev. & Mean & Std. Dev. & Mean & Std. Dev.  & Mean & Std. Dev. & Mean & Std. Dev. &  Mean & Std. Dev. \\
				\hline
				$f_{21}$ &4.00E+02 &0.00E+00 &3.94E+02 &4.09E+01  &3.90E+02 &4.96E+01 &\textbf{2.35E+02} &\textbf{9.04E+01} &3.69E+02 &7.28E+01  &3.96E+02 &2.78E+01\\
				$f_{22}$ &3.15E+02 &1.06E+02 &5.45E+01 &2.32E+01  &2.95E+01 &2.39E+01 &1.73E+01 &2.60E+01 &\textbf{1.26E+01} &\textbf{1.52E+01}  &2.02E+01 &3.10E+01\\
				$f_{23}$ &1.13E+03 &1.81E+02 &1.08E+03 &1.43E+02  &\textbf{6.04E+02} &\textbf{2.02E+02} &7.66E+02 &2.62E+02 &6.52E+02 &2.27E+02  &6.28E+02 &1.99E+02\\
				$f_{24}$ &\textbf{1.52E+02} &\textbf{3.28E+01} &1.72E+02 &3.80E+01  &1.74E+02 &3.79E+01 &1.87E+02 &3.80E+01 &1.70E+02 &4.29E+01  &1.88E+02 &3.31E+01\\
				$f_{25}$ &1.99E+02 &5.84E+00 &1.94E+02 &1.76E+01  &1.96E+02 &1.65E+01 &2.01E+02 &7.87E+00 &\textbf{1.92E+02} &\textbf{2.60E+01}  &1.93E+02 &2.46E+01\\
				$f_{26}$ &1.17E+02 &1.23E+01 &1.14E+02 &1.26E+01  &1.07E+02 &1.33E+01 &1.16E+02 &1.80E+01 &1.10E+02 &1.34E+01  &\textbf{1.07E+02} &\textbf{6.15E+00} \\
				$f_{27}$ &3.08E+02 &2.22E+01 &3.01E+02 &3.48E+00  &\textbf{3.00E+02} &\textbf{3.33E-02} &3.31E+02 &4.85E+01 &3.10E+02 &2.97E+01  &3.06E+02 &2.35E+01\\
				$f_{28}$ &2.97E+02 &1.91E+01 &2.97E+02 &2.09E+01  &2.96E+02 &2.77E+01 &\textbf{2.25E+02} &\textbf{9.67E+01} &2.65E+02 &7.62E+01  &2.76E+02 &6.44E+01\\
				\hline
		\end{tabular}}
	\end{center}
\end{table*}

\begin{table*}[htbp]
	\caption{The results of last 8 problems of CEC'13 with different cell size for $n=30$ at termination}
	\label{13_30D_CELLS_MAXFE}
	\begin{center}
		\begin{tabular}{c|cc|cc|cc|cc|cc}
			\hline
			{$N$} &\multicolumn{2}{c|}{100} &\multicolumn{2}{c|}{100} &\multicolumn{2}{c|}{100} &\multicolumn{2}{c|}{150} &\multicolumn{2}{c}{200} \\
			{cell size} &\multicolumn{2}{c|}{2000} &\multicolumn{2}{c|}{3000} &\multicolumn{2}{c|}{3500} &\multicolumn{2}{c|}{3000} &\multicolumn{2}{c}{1500} \\
			\hline
			{} & Mean & Std. Dev. & Mean & Std. Dev. &  Mean & Std. Dev. &  Mean & Std. Dev. & Mean & Std. Dev. \\
			\hline
			$f_{21}$ &3.36E+02 &9.03E+01 &3.34E+02 &9.21E+01  &\textbf{3.08E+02} &\textbf{6.18E+01} &3.42E+02 &8.43E+01 &3.21E+02 &5.01E+01\\
			$f_{22}$ &\textbf{1.33E+02} &\textbf{1.29E+01} &3.04E+02 &8.11E+01  &3.69E+03 &2.74E+02 &5.04E+02 &1.23E+02 &3.13E+03 &1.62E+02\\
			$f_{23}$ &\textbf{3.43E+03} &\textbf{4.80E+02} &4.28E+03 &9.20E+02  &7.11E+03 &3.02E+02 &4.14E+03 &7.97E+02 &6.88E+03 &3.56E+02\\
			$f_{24}$ &2.09E+02 &4.68E+00 &2.07E+02 &3.75E+00  &2.07E+02 &3.71E+00 &2.04E+02 &2.46E+00 &\textbf{2.02E+02} &\textbf{2.01E+00}\\
			$f_{25}$ &2.58E+02 &1.32E+01 &\textbf{2.46E+02} &\textbf{3.09E+01}  &2.50E+02 &2.86E+01 &2.73E+02 &2.37E+01 &2.73E+02 &3.26E+01\\
			$f_{26}$ &2.00E+02 &3.60E-03 &2.00E+02 &2.77E-03  &2.00E+02 &1.30E-02 &\textbf{2.00E+02} &\textbf{1.98E-03} &2.00E+02 &1.21E-02\\
			$f_{27}$ &4.51E+02 &9.55E+01 &3.69E+02 &3.38E+01  &3.49E+02 &2.65E+01 &5.79E+02 &1.25E+02 &\textbf{3.31E+02} &\textbf{2.29E+01}\\
			$f_{28}$ &3.00E+02 &0.00E+00 &3.00E+02 &0.00E+00  &\textbf{2.99E+02} &\textbf{4.33E+00} &3.00E+02 &0.00E+00 &3.00E+02 &0.00E+00\\
			\hline
		\end{tabular}
	\end{center}
\end{table*}

Six agents with different number of neurons are learned on 10-D functions. A set of neuron sizes, from 500 to 3000 with an interval of 500, is studied when the population size is fixed as 50. The obtained results are summarized in Table~\ref{13_10D_CELLS_MAXFE}. 

From Table~\ref{13_10D_CELLS_MAXFE}, we see that 1) the performance of LDE differs w.r.t. the size of neurons; 2) for different functions, the best result is obtained by taking different neuron size; and 3) a higher number of neurons does not always mean better performance.

Generally speaking, the population size ought to be increased for problems with larger dimensions. To see the effect of population size, we carry out experiments to learn five controllers that are with different population and neuron sizes on the same CEC'13 training functions (i.e. $f_1-f_{20}$ with 30-D). The performance of the learned controller is again tested on $f_{21}-f_{28}$ with 30-D. 

Table~\ref{13_30D_CELLS_MAXFE} lists the comparison results of the five designed controllers. From the table, it is observed that the neuron size takes the same effects as those in 10-D case. Further, it can be seen that the population and neuron size together have a very complex effect on the performance of the learned controller.

\section{Related Work}\label{rw}

In this paper, RL is used as the main technique to learn on how to adaptively control the algorithmic parameters from optimization experiences. To the best of our knowledge, there is no related work on controlling the DE parameters by learning from optimization experiences. However, we found some works on controlling the parameters of genetic algorithms (GAs). These works relate to our approach but with significant differences.

In~\cite{Eiben07}, four control parameters (including crossover rate, mutation rate, tournament proportion and population size) of a GA are dynamically regulated with the help of the reinforcement learning. The learning algorithm is a mix of Q-learning and SARSA which involves maintaining a discrete table of state-action pairs. In~\cite{Eiben07}, information along the GA's search procedure is extracted as the state. Two RL algorithms switch in a pre-defined frequency to find a new action (i.e. control parameter value). The work shows that the RL-enhanced GA outperforms a steady-state GA in terms of fitness and success rate.

In~\cite{Sakurai10}, the Q-learning algorithm is applied to choose a suitable reproduction operator which can generate a promising individual in a short time. The authors propose a new reward function incorporating GA's multi-point search feature and the time complexity of recombination operators. Further, the action-value function is updated after generating all individuals. Similarly, in~\cite{Buzdalova14}, the Q-learning is also used to adaptively select reproduction operators. But the chosen operator is applied to the whole population. This method is shown empirically that it tends to avoid obstructive operators and thus solve the problems more efficiently than random selection.

In~\cite{Karafotias14}, a universal controller by using RL is found to be able to adapt to any existing EA and to adjust any parameter involved. In their method, a set of observables (considered as states) is fed to a binary decision tree consisting of only one root node for representing a universal state. SARSA~\cite{rlbook} is carried out to update the state-action value. It is shown that the RL-enhanced controller exhibits superiority over two benchmark controllers on most common complex problems.

Here we would like to point out the significant differences between the proposed approach and the aforementioned RL based approaches. First, the RL methods, such as Q-learning or SARSA used in existing approaches are developed for MDP with the discrete state and action. Second, existing parameter controllers are not learnt from optimization experiences, but are updated based on the online information obtained from the search procedure during a single run for a single test problem. The main idea behind the existing study is the same as in the DE parameter control methods reviewed in the introduction. They all try to use information obtained online to update the control parameters. Third, different from RL which aims to learn an agent with a converged optimal policy, the policy derived from the state-action pairs in existing study is not necessarily convergent or even stable. However, the proposed approach in this paper can learn from the extraneous information for a stable policy. 

The only work that applies an idea similar to our approach is the DE-DDQN~\cite{Sharma19}, in which a set of mutation operators is adaptively selected based on the learning from optimization experiences over a set of training functions. In DE-DDQN, double deep Q learning is applied for the selection. Various features are defined as states and taken as input to the deep neural network at each generation. 

\section{conclusion}\label{con}

This paper proposed a new adaptive parameter controller by learning from the optimization experiences of a set of training functions. The adaptive parameter control problem was modeled as an MDP. A recurrent neural network, called LSTM, was employed as the parameter controller. The reinforcement learning algorithm, policy gradient, was used to learn the parameters of the LSTM. The learned controller was embedded within a DE for new test problem optimization. In the experiments, functions in the CEC'13 test suite were used in training. After training, the trained agent was studied on the CEC'13 and CEC'17 test suites in comparison with some well-known DE algorithms and a state-of-the-art evolutionary algorithm. The experimental results showed that the learned DE was very competitive to the compared algorithms which indicated the effectiveness of the proposed controller. 

From our experimental study, we find that training the parameter controller for 30D problems is rather difficult in terms of the computational resources. Particularly,  the CPU/GPU time used in the training process is considerable. Further, as the number of dimension increases, it is expected that there will be an increasing need for training time and powerful computing devices. It is also hard to choose the training functions to make the training stable. Moreover, there has no theoretical foundation or practical principles on deciding the cell size in the employed neural network. Another disadvantage of the learned algorithm is that its time complexity is greater than the compared algorithms. 
    
Note that the training and test functions share similar features since they are all constructed by using the same basic functions. As a result, its performance over unrelated functions is not predictable, may be limited on totally different set of functions, such as real-world problems. A possible way to improve the applicability of LDE maybe is to use new learning techniques or incorporate existing DE techniques in the LDE. 

In the future, we plan to improve the performance of the LDE in a number of ways, such as using different statistics ${\cal U}^t$, adopting different neural networks, considering different output of the neural network, and others. Further, we intend to apply the LDE on some real-world optimization and engineering problems. We also intend to study on the use of reinforcement learning for adaptive mutation/crossover strategy, on the learning for hyper-parameters of state-of-the-art evolutionary algorithms,  and on the learning for meta-heuristics for combinatorial optimization problems.

\bibliographystyle{IEEEtran}
\bibliography{LDE}
\end{document}